\batchmode
\makeatletter
\def\input@path{{\string"C:/Users/Yuan Zhou/Dropbox/YuanHyperspectral/document/GMM_SantaBarbara/\string"}}
\makeatother
\documentclass[twocolumn,english]{IEEEtran}
\usepackage[T1]{fontenc}
\usepackage[latin9]{inputenc}
\usepackage{color}
\usepackage{babel}
\usepackage{array}
\usepackage{rotating}
\usepackage{float}
\usepackage{mathtools}
\usepackage{multirow}
\usepackage{amsmath}
\usepackage{amssymb}
\usepackage{graphicx}
\usepackage[unicode=true]
 {hyperref}

\makeatletter

\providecommand{\tabularnewline}{\\}
\floatstyle{ruled}
\newfloat{algorithm}{tbp}{loa}
\providecommand{\algorithmname}{Algorithm}
\floatname{algorithm}{\protect\algorithmname}

 \let\oldforeign@language\foreign@language
 \DeclareRobustCommand{\foreign@language}[1]{%
   \lowercase{\oldforeign@language{#1}}}

\usepackage[caption=false,font=footnotesize]{subfig}
\usepackage{threeparttable}

\makeatother

\begin{document}

\title{Unmixing urban hyperspectral imagery with a Gaussian mixture model
on endmember variability}

\author{Yuan~Zhou,~\IEEEmembership{Student Member,~IEEE,} Erin~B.~Wetherley,
and~Paul~D.~Gader,~\IEEEmembership{Fellow,~IEEE}\thanks{Y. Zhou and P. D. Gader are with the Department of Computer and Information
Science and Engineering, University of Florida, Gainesville, FL, USA.
E-mail: \protect\href{mailto:{yuan,pgader}@cise.ufl.edu}{{yuan,pgader}@cise.ufl.edu}.}\thanks{E. B. Wetherley is with the Department of Geography, University of
California Santa Barbara, Santa Barbara, CA, USA. E-mail: \protect\href{mailto:wetherley@umail.ucsb.edu}{wetherley@umail.ucsb.edu}.}}

\markboth{}{Y. Zhou \MakeLowercase{\emph{et al.}}: Unmixing urban hyperspectral
imagery with a Gaussian mixture model on endmember variability}

\IEEEpubid{0000\textendash 0000/00\$00.00~\copyright~2018 IEEE}
\maketitle
\begin{abstract}
Spectral unmixing given a library of endmember spectra can be achieved
by multiple endmember spectral mixture analysis (MESMA), which tries
to find the optimal combination of endmember spectra for each pixel
by iteratively examining each endmember combination. However, as library
size grows, computational complexity increases which often necessitates
a laborious and heuristic library reduction method. In this paper,
we model a pixel as a linear combination of endmembers sampled from
probability distributions of Gaussian mixture models (GMM). The parameters
of the GMM distributions are estimated using spectral libraries. Abundances
are estimated based on the distribution parameters. The advantage
of this algorithm is that the model size grows very slowly as a function
of the library size.

To validate this method, we used data collected by the AVIRIS sensor
over the Santa Barbara region: two 16 m spatial resolution and two
4 m spatial resolution images. 64 validated regions of interest (ROI)
(180 m by 180 m) were used to assess estimate accuracy. Ground truth
was obtained using 1 m images leading to the following 6 classes:
turfgrass, non-photosynthetic vegetation (NPV), paved, roof, soil,
and tree. Spectral libraries were built by manually identifying and
extracting pure spectra from both resolution images, resulting in
3,287 spectra at 16 m and 15,426 spectra at 4 m. We then unmixed ROIs
of each resolution using the following unmixing algorithms: the set-based
algorithms MESMA and AAM, and the distribution-based algorithms GMM,
NCM, and BCM. The original libraries were used for the distribution-based
algorithms whereas set-based methods required a sophisticated reduction
method, resulting in reduced libraries of 61 spectra at 16 m and 95
spectra at 4 m. The results show that GMM performs best among the
distribution-based methods, producing comparable accuracy to MESMA,
and may be more robust across datasets.
\end{abstract}


\section{Introduction}

\IEEEpubidadjcol

\IEEEPARstart{H}{yperspectral} images have important applications
in astronomy, agriculture, geoscience, surveillance (such as object
identification), material identification, and detecting processes
\cite{vane1993airborne}. Because limited photons enter the sensor
when collecting narrow bandwidth channels from a high altitude, the
spatial resolution of hyperspectral image is usually very coarse,
i.e. a pixel may correspond to a region with a diameter of several
meters. Hence, multiple materials may exist in this region and contribute
to the measured pixel spectrum, also known as a mixed pixel \cite{small2001estimation}.
One important problem in hyperspectral imagery is to decompose mixed
pixels to identify the constituting materials (\emph{endmember}) and
their proportions (\emph{abundance}) that form the pixel spectrum.

The most common model that relates endmembers and abundances to a
pixel is the \emph{linear mixing model} (LMM), which assumes that
the reflectance measured within each pixel is a unique linear combination
of the reflectances of each sub-pixel endmember, weighted by its abundance,
plus some noise \cite{settle1993linear}. The intuition behind this
model is that the fractional area of a material determines its representation
in the measured signal. However, when unmixing a hyperspectral image
with LMM, we usually encounter an additional problem that spectral
reflectance for identical materials are often different. For example,
asphalt spectra can vary significantly based on age, shadowing, and
composite materials \cite{herold2005spectral}. This is sometimes
called endmember variability \cite{zare2014endmember,somers2011endmember}.

\IEEEpubidadjcol

Several factors can contribute to endmember variability, including
both extrinsic factors and intrinsic factors. The most significant
extrinsic factor is illumination. When solar incidence and emergence
angles are different for a surface, the observed signal will be different
\cite{meister2000brdf}. Material angle matters as well, for example
roofs can be present at a variety of angles relative to incoming solar
radiation, producing different spectral signatures for one material.
Atmospheric condition can be another extrinsic factor affecting reflectance,
however this is usually corrected during image processing. Measurement
scale represents an important intrinsic factor. Objects or materials
that may be considered \textquotedblleft pure\textquotedblright{}
may in reality be composed of materials at smaller scales with varying
reflectances \cite{wentz2012synthesizing}. For example, a tree canopy
can be considered a single, pure endmember, however this ignores the
spectral variety of tree leaves, bark, branches, and substrate that
composes a single tree pixel \cite{roberts2004spectral}. Similarly,
soils are composed of particles with different shapes, sizes, and
chemical composition \cite{somers2011endmember}. The larger scale
we use to define an endmember, the larger intrinsic variability we
may expect from its spectra. For example, trees and turfgrass can
be defined as individual endmembers, however if we we wish to define
a class of green vegetation comprised of both turfgrass and tree,
its variability will not be less than the component endmember.

Considering endmember variability, we can generalize the LMM to the
following equation:

\begin{equation}
\mathbf{y}_{n}=\sum_{j=1}^{M}\mathbf{m}_{nj}\alpha_{nj}+\mathbf{n}_{n},\,n=1,\dots,N\label{eq:LMM_n}
\end{equation}
where $\mathbf{y}_{n}\in\mathbb{R}^{B}$ is the spectrum of the $n$th
pixel in the image, $B$ is the number of bands, $N$ is the number
of pixels, $M$ is the number of endmembers. $\mathbf{m}_{nj}\in\mathbb{R}^{B}$
is the $j$th endmember for the $n$th pixel. $\alpha_{nj}\in\mathbb{R}$
is the abundance that usually satisfies the positivity and sum-to-one
constraints, i.e. $\alpha_{nj}\ge0,\,\sum_{j}\alpha_{nj}=1$. Finally,
we have some additive noise $\mathbf{n}_{n}$.

When it comes to unmixing in terms of \eqref{eq:LMM_n}, we are referring
to retrieving $\left\{ \mathbf{m}_{nj},\alpha_{nj}\right\} $ from
$\left\{ \mathbf{y}_{n}\right\} $, or $\left\{ \alpha_{nj}\right\} $
from $\left\{ \mathbf{y}_{n}\right\} $ and a library of endmember
spectra. The former is sometimes called \emph{unsupervised} \emph{unmixing},
and because it is undetermined this can be a difficult problem. Studies
that have worked to solve unsupervised unmixing usually require several
assumptions, such as spatial smoothness of the abundances and the
existence of contiguous pure pixels \cite{halimi2015unsupervised,drumetz2016blind,zhou2018gmmJournal}.
The latter is called \emph{supervised unmixing} and depends on a library
of known endmember spectra. If the library is small enough to easily
enumerate all possible spectral combinations, the task can be trivial.
However, applying this scheme on larger libraries becomes computationally
inefficient. This is the problem we are addressing in this study.

Previous studies that have worked to solve this problem have used
methods that can be categorized as \emph{set-based} or \emph{distribution-based}
\cite{zare2014endmember}. Set-based methods treat the endmember library
as an unordered set and try to pick the best combination of endmembers
to model each pixel. A widely used set-based method is multiple endmember
spectral mixture analysis (MESMA) \cite{roberts1998mapping}. The
general idea of MESMA is to test every endmember combination and select
the one with the smallest error within set thresholds that limit pixel
complexity. There are many variations to MESMA. In multiple-endmember
linear spectral unmixing model (MELSUM), the solution for abundances
is obtained from directly solving the linear equations and discarding
the negative values \cite{combe2008analysis}. In automatic Monte
Carlo unmixing (AutoMCU), pixels are unmixed using multiple sets of
random combinations, with the mean fractional values assigned as abundances
\cite{asner2000biogeophysical,asner2002spectral}. In alternate angle
minimization (AAM), projection is iteratively used to find the spectrum
index of one endmember given the other endmembers fixed. Besides MESMA
variants, there is sparse unmixing that used the full spectral library
with a sparsity constraint on the abundances forcing them having only
a few nonzero elements \cite{castrodad2011learning}.

Contrary to set-based methods, distribution-based methods assume that
the endmembers for each pixel are sampled from probability distributions,
hence the linear combinations of these endmembers (pixels) also follow
some distribution. It works by modeling the spectral library as statistical
distributions, extracting parameters to describe these distributions,
and unmixing the pixels based on the distribution parameters. The
most widely used distribution is Gaussian, and its application for
spectral unmixing is known as the normal compositional model (NCM)
\cite{eches2010bayesian,eches2010estimating,halimi2015unsupervised,stein2003application,zare2010pce,zhangpso}.
The popularity of NCM comes from the fact that a linear combination
of Gaussian random variables is also a Gaussian random variable whose
mean and covariance matrix are linear combinations from the endmember
means and covariance matrices. Hence, the resulting probability density
function of the pixels has a simple analytical form. Fitting the actual
pixel values to the pixel distribution, the abundances can be solved
by several techniques, such as expectation maximization \cite{stein2003application},
sampling methods \cite{eches2010bayesian,eches2010estimating,halimi2015unsupervised},
and particle swarm optimization \cite{zhangpso}.

Following this philosophy, some have worked to extend the idea to
distributions beyond Gaussian. In \cite{du2014spatial}, the authors
propose Beta distributions to model the spectral library. The benefit
is that Beta distributions have a domain in the range 0 \textendash{}
1, so are more suitable for the reflectance range, and the actual
library may have a skewed mode in the distribution. In \cite{zhougaussian,zhou2018gmmJournal},
the idea is further extended to use Gaussian mixture models (GMM)
for distributions. The rationale comes from the observation that library
endmembers may have multiple modes, whose shape cannot be represented
by a simple Gaussian or Beta distribution. Since GMM is more flexible,
it can approximate any distribution found in the library.

\subsection{Our contribution}

Many unmixing studies are not well evaluated in presence of ground
truth. Commonly used hyperspectral datasets include Pavia University,
Indian Pines, Cuprite, Mississippi Gulfport, etc., which are not validated
with ground truth endmembers and abundances. Hence, the primary method
for evaluating their results include:
\begin{enumerate}
\item Compare the estimated endmembers with spectra in the USGS spectra
library (e.g. in Cuprite dataset) \cite{nascimento2005vertex,lu2013manifold}.
\item Compare the estimated abundances with assumed segmentation maps of
pure materials (e.g. in Indian Pines, Pavia University, Gulfport datasets)
\cite{zare2010pce,zare2013piecewise}.
\item Calculate the reconstruction error of estimated endmembers and abundances
and assume that a lower reconstruction error implies a better result
\cite{halimi2015unsupervised,drumetz2016blind}.
\end{enumerate}
Each of these methods can be problematic. First, different conditions
(sensor, atmosphere, light source) during data collection will affect
measured reflectances, making library comparison less ideal. Second,
high spatial resolution hyperspectral images are primarily composed
of pure pixels, and segmentation like abundance maps do not necessarily
indicate good unmixing capability for mixed pixels. Third, reconstruction
error is more related to model complexity than unmixing accuracy since
small reconstruction error could be achieved by overfitting \cite{murphy2012machine}.

Moreover, these datasets are not comprehensive with respect to spatial
scales, scene diversity and generalization. For example, the Pavia
University and Gulfport datasets have about 1 m spatial resolution
in which most are pure pixels. Also, they are focused on only a few
urban sites, which contain mostly man-made materials with segmentation
like abundance distribution. Developing unmixing algorithms on them
will have a bias on forcing smooth and sparse abundance maps. Hence,
it is unknown if the algorithms validated on these datasets can be
applied to datasets with generalized scenarios.

In this work, we introduce a supervised unmixing algorithm based on
modeling endmember variability by GMM distributions, and compare several
set-based and distribution-based algorithms with a highly validated,
comprehensive dataset of 128 images with different spatial scales.
The algorithm was first introduced in \cite{zhou2018gmmJournal,zhougaussian}
and we modified it for this application. The dataset was developed
in \cite{wetherley2017mapping} but only used for evaluating MESMA.
It contains two types of images, one with about 16 m pixel size, the
other with about 4 m pixel size. It covers a wide range of landcover,
including various kinds of road, roof, vegetation, and soil. Validation
abundances were obtained by classifying high resolution images corresponding
to the hyperspectral images. Unlike MESMA, which requires a small
and well-curated spectral library, the GMM algorithm uses the original
source library without modification to unmix fractions using inferred
parameters from the library.

\section{Dataset\label{sec:Dataset}}

We used two low-resolution images (16 m) and two high-resolution images
(4 m) in this study. The low-resolution images were collected by the
Airborne Visible/Infrared Imaging Spectrometer (AVIRIS) \cite{vane1993airborne}
over Santa Barbara, CA, on August 29, 2014. The spatial resolutions
are 15.6 m/pixel and 15.8 m/pixel. The spectral range measures wavelengths
from 380 \textendash{} 2500 nm with 224 bands of approximately 10
nm bandwidth. High-resolution images were collected by AVIRIS-Next
Generation with 3.9 m/pixel and 3.6 m/pixel spatial resolutions. The
spectral resolution is also higher, recording 432 bands of about 5
- 6 nm bandwidth across a similar spectral range as the 16 m dataset.
We spectrally resampled the AVIRIS-Next Generation imagery to 224
bands to produce an image with identical spectral parameters to the
16 m AVIRIS image. We also removed certain bands from analysis due
to atmospheric interference, reducing the number of bands to 164.
Initial image processing was conducted by the Jet Propulsion Laboratory,
with additional processing in the lab to reduce the effects of elevation
change on pixel location. 

The study area includes the cities of Santa Barbara and Goleta as
well as the land between them, near the California coast. Urban composition
is typical of the southwestern United States, including man-made materials
such as asphalt, concrete, metal, gravel, and brick, as well as vegetation
in the forms of turfgrass, various tree species, and large areas of
undeveloped land covered in senesced vegetation \cite{roberts2012synergies}.

\subsection{Validation Polygons}

\begin{figure*}
\begin{centering}
\includegraphics[width=18cm]{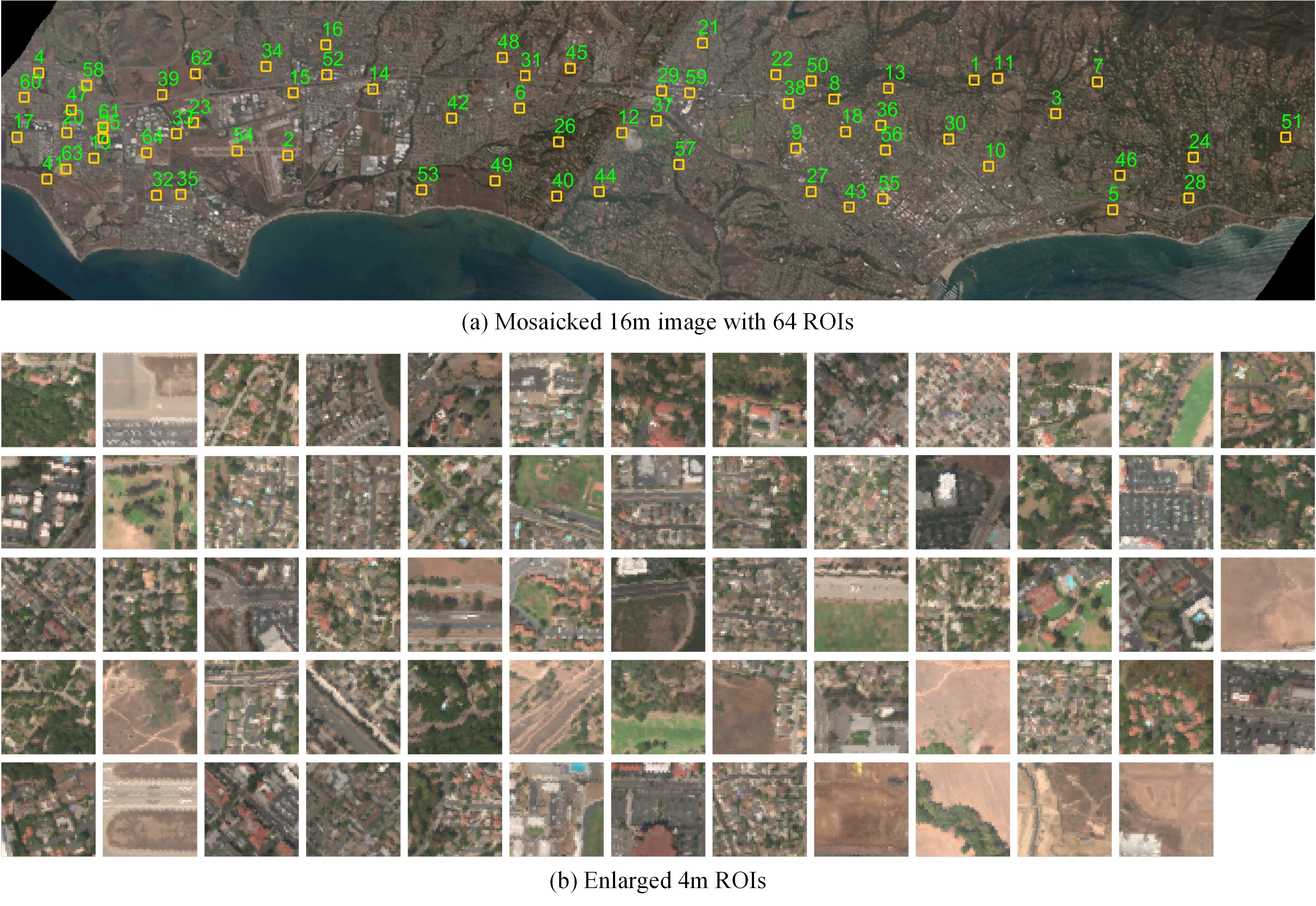}
\par\end{centering}
\caption[Validation polygons on the site (a) and all 4 m ROI images (b).]{Validation polygons on the site (a) and all 4 m ROI images (b). The
two 16 m images are mosaicked by geographic coordinates. }

\label{fig:validation_polygons}
\end{figure*}

We produced 64 polygons that represented the variety of landcover
within the study area. Each polygon was 180 m by 180 m in size, or
11-12 pixels wide in the 16 m images (46 or 50 pixels wide in the
4 m images). Validation polygons were randomly distributed across
the area with a minimum distance 400 m. If a polygon contained large
areas of open water or an undetermined material, it was discarded
and a new polygon randomly generated. Cover was determined within
each polygon using a 1 m NAIP high-resolution image. We used a combination
of image segmentation, using ECognition, and manual adjustments to
classify the cover within each polygon as turf, tree, paved, roof,
soil, or non-photosynthetic vegetation (NPV). Cover was further confirmed
by visually inspecting August 2014 Google Earth imagery. Fig~\ref{fig:validation_polygons}
displays all polygons as they appear in the 16 m images. 

\begin{figure}
\begin{centering}
\includegraphics[width=7.5cm]{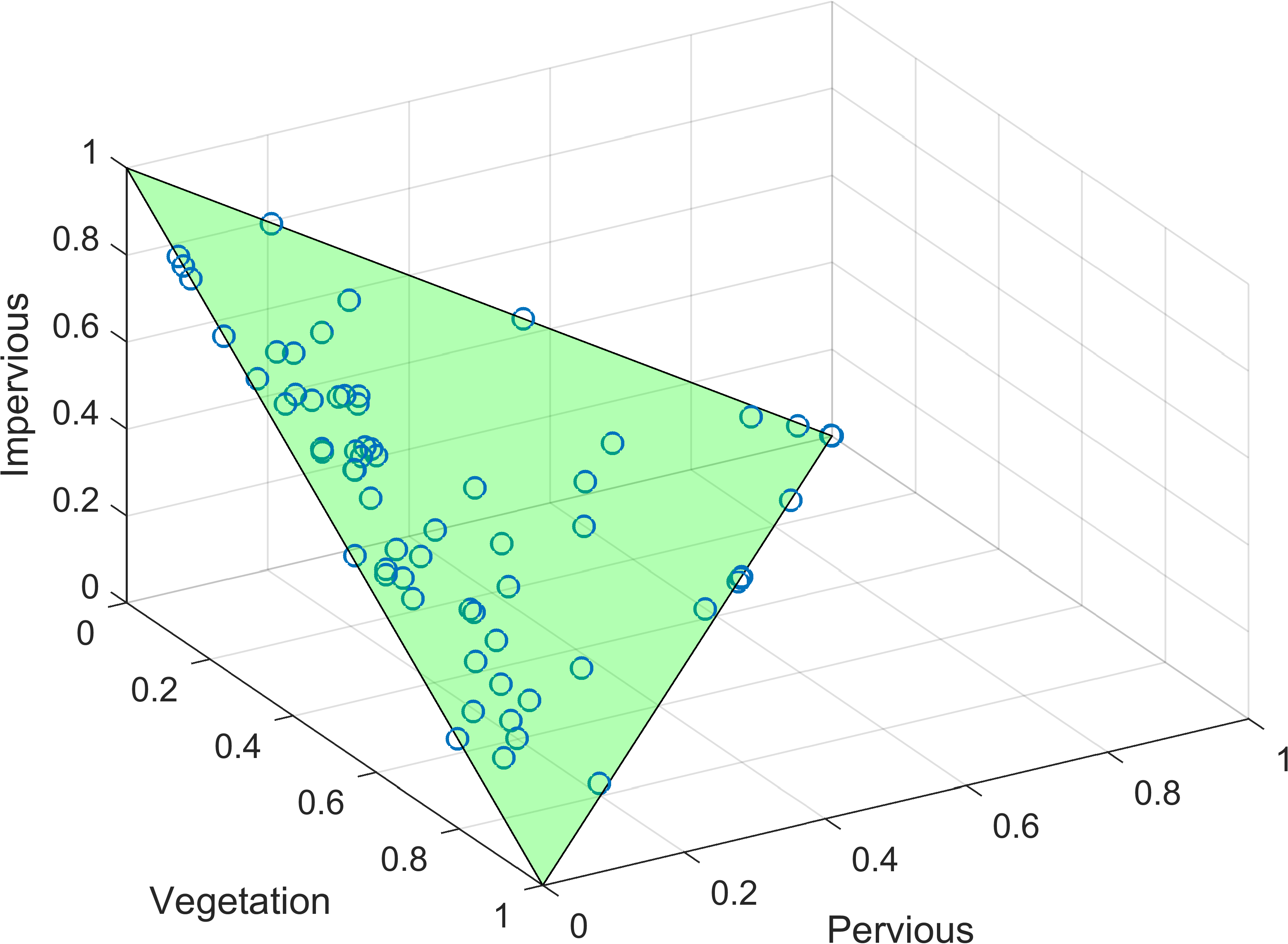}
\par\end{centering}
\caption{Scatter plot of ground truth total abundances in terms of 3 categories,
green vegetation (turfgrass and tree), pervious (NPV and soil), and
impervious (paved and roof). Most of them lie on the plane, which
corresponds with the selection of ROIs where almost all the pixels
fall into the 6 endmember classes.}

\label{fig:abund_gt_scatter}
\end{figure}

Fig.~\ref{fig:abund_gt_scatter} shows a scatter plot of the 64 ground
truth abundances when the 6 endmember classes are merged to 3 categories
of vegetation, impervious, and non-vegetated pervious. Most polygons
are dominated by a mixture of impervious and vegetation materials.
To improve the representation of less common mixtures in the scene,
we added 5 polygons with high proportions of soil. 

\subsection{Library Building}

We produced 240 polygons across the 4 m scene to extract pure spectra
and build the full spectral libraries. The polygons were intended
to capture class material variability as much as possible, and so
included multiple roof types, asphalt, concrete, trees, turfgrass,
soil, and NPV, as well as less common materials like rubber, solar
panels, tennis courts, and plastic tarps. These materials were then
grouped into one of our 6 endmember classes: turfgrass, NPV, paved,
roof, soil, and tree.

The same polygons were used to extract spectra from the 16 m imagery,
with necessary modifications as described in \cite{wetherley2017mapping}.
Together, we produced a library of 16 m spectra and a library of 4
m spectra. After removing duplicate spectra, the final 16 m library
was comprised of 3,287 spectra and the 4 m library contained 15,426
spectra. 

\begin{figure*}
\begin{centering}
\includegraphics[width=16.5cm]{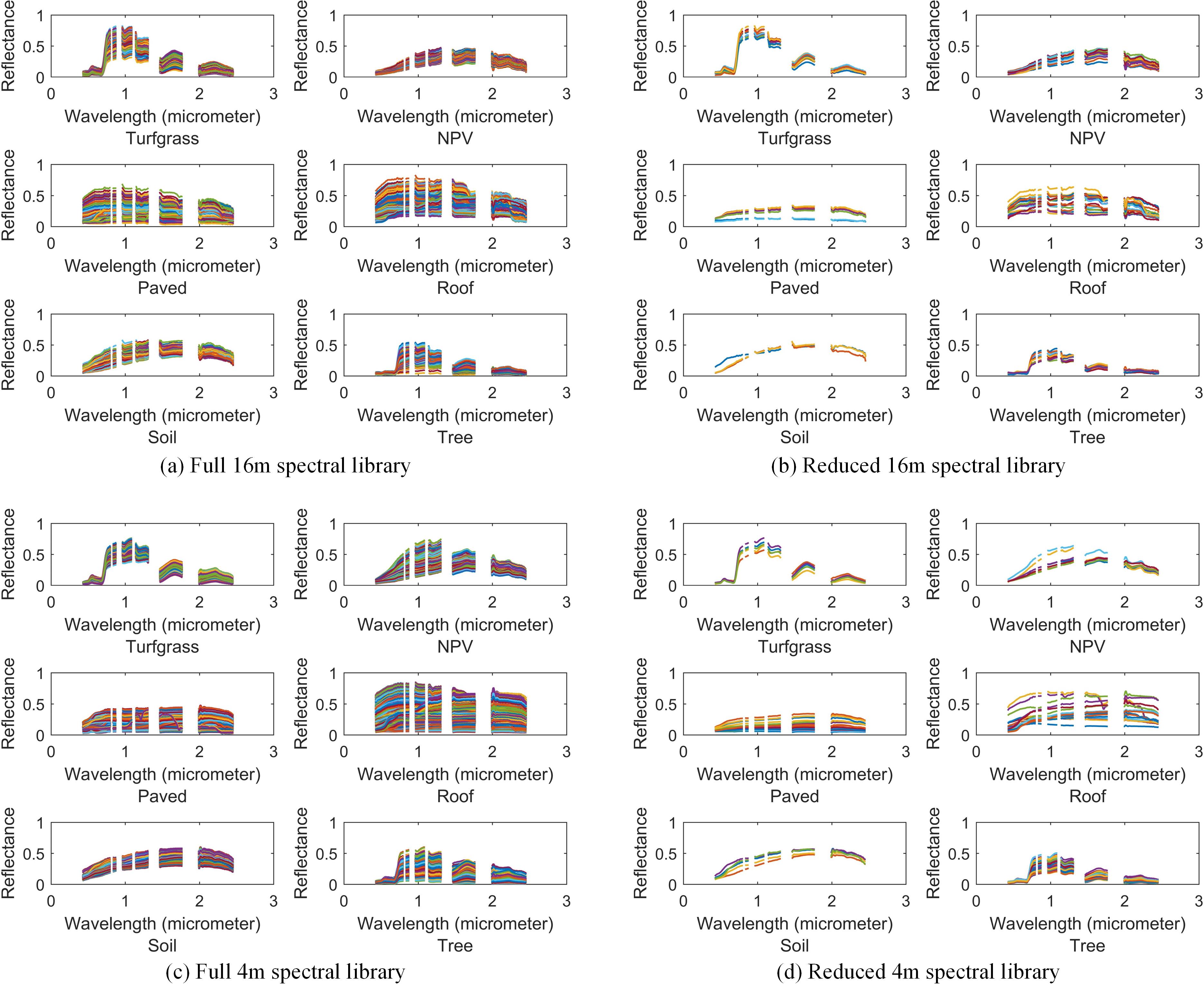}
\par\end{centering}
\caption[Original and reduced spectral libraries.]{Original and reduced spectral libraries. The numbers of spectra in
each category are shown in Table~\ref{table:num_spectra}.}

\label{fig:spectral_library}
\end{figure*}

These full spectral libraries were used to train the parameters of
distribution-based algorithms. However, they were too large to be
used by MESMA, and required reduction. We performed reduction in two
steps. First, iterative endmember selection (IES) \cite{schaaf2011mapping}
was used to automatically select a subset of spectra that represented
the larger library. This is achieved iteratively, by gradually selecting
the most representative spectra and evaluating their representativeness
using a kappa coefficient. IES reduced the 16 m and 4 m library sizes
to 226 and 187, respectively. Libraries were further reduced using
iterative classification reduction (ICR), which uses MESMA as a classifier
to quickly identify and remove spectra that tend to map materials
incorrectly \cite{wetherley2017mapping}. This reduced the libraries
to a final size of 61 for 16 m images and 95 for 4 m images. The spectra
for each endmember class for all the cases are plotted in Fig~\ref{fig:spectral_library},
and their numbers are shown in Table~\ref{table:num_spectra}.

\begin{table}
\caption{Number of spectra for each endmember class in the libraries}

\begin{centering}
\begin{tabular}{|c|c|c|c|c|}
\hline 
 &
\multicolumn{2}{c|}{16 m} &
\multicolumn{2}{c|}{4 m}\tabularnewline
\hline 
\hline 
 &
Full &
Reduced &
Full &
Reduced\tabularnewline
\hline 
Turfgrass &
537 &
10 &
1468 &
5\tabularnewline
\hline 
NPV &
884 &
14 &
3465 &
7\tabularnewline
\hline 
Paved &
299 &
6 &
2902 &
17\tabularnewline
\hline 
Roof &
435 &
17 &
2941 &
16\tabularnewline
\hline 
Soil &
262 &
3 &
1442 &
5\tabularnewline
\hline 
Tree &
870 &
11 &
3208 &
45\tabularnewline
\hline 
Total &
3287 &
61 &
15426 &
95\tabularnewline
\hline 
\end{tabular}
\par\end{centering}
\label{table:num_spectra}
\end{table}

\section{Method}

\subsection{The Gaussian Mixture Model for Unmixing\label{subsec:The-Gaussian-mixture}}

\begin{figure*}
\begin{centering}
\includegraphics[width=16cm]{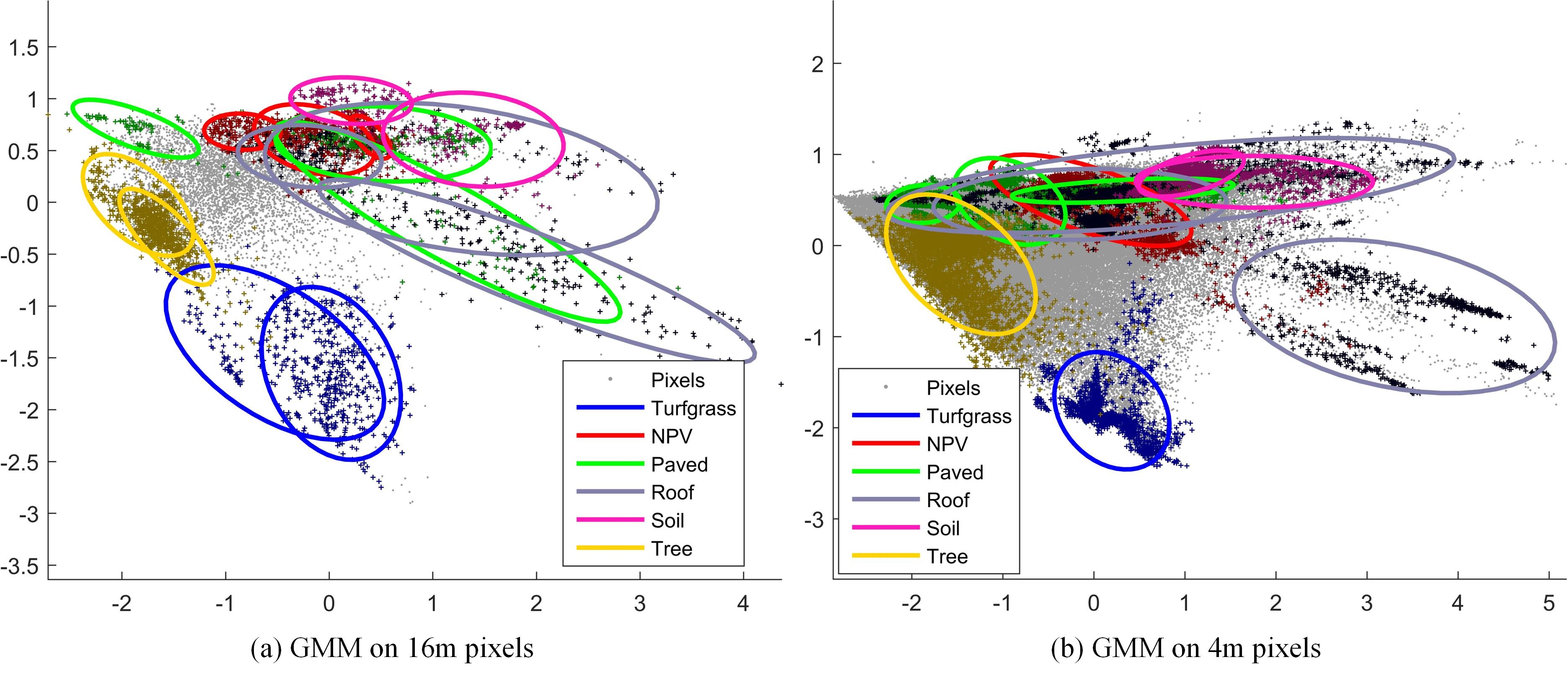}
\par\end{centering}
\caption[Scatter plot of GMM components on the pixels and library spectra.]{Scatter plot of GMM components on the pixels and library spectra.
The projection is determined by performing PCA on all the spectra
in the library. The pixels of 64 images for each scale are combined
and denoted by gray dots. The colored dots show the spectra in the
library for each endmember class. The ellipses denote the Gaussian
components.}

\label{fig:gmm_scatter}
\end{figure*}

\begin{figure*}
\begin{centering}
\includegraphics[width=16cm]{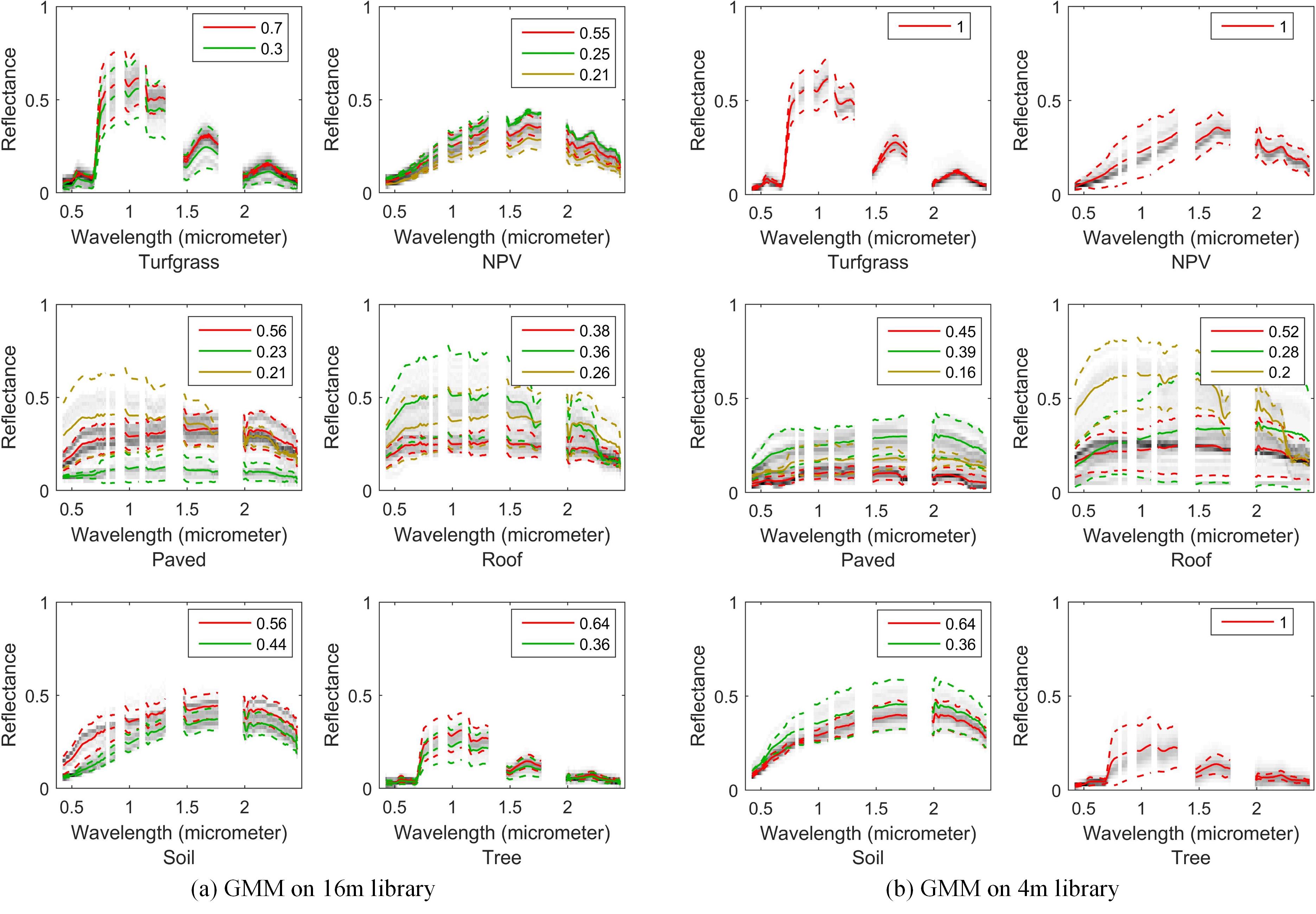}
\par\end{centering}
\caption[Wavelength-reflectance plot of GMM components on the library spectra.]{Wavelength-reflectance plot of GMM components on the library spectra.
The spectra are put into 2-dimensional bins of wavelength-reflectance
to form a histogram shown as gray scale background images. The center
of each Gaussian component is shown as solid curve. The center plus
(minus) twice the square root of the largest eigenvalue with its corresponding
eigenvector is shown as a dashed curve, which indicates the major
variation pattern of a Gaussian component. The prior probabilities
are shown in the legends.}

\label{fig:gmm_wv_refl}
\end{figure*}

Here we briefly introduce the GMM based unmixing \cite{zhou2018gmmJournal},
which is a generative model that models a distribution on the input
space \cite{bishop2006pattern}. Suppose we have $M$ endmember classes,
each has numerous spectra in the library. A pixel can be assumed to
be generated by randomly picking one spectrum for each endmember,
and linearly mixing them based on some abundances. In this way, if
we use a probability density function to represent the spectral distribution,
the actual endmembers can be assumed to be sampled from this distribution.
Suppose the $j$th endmember for the $n$th pixel is sampled from
a distribution modeled by GMM
\[
p\left(\mathbf{m}_{nj}\vert\boldsymbol{\Theta}\right)=\sum_{k=1}^{K_{j}}\pi_{jk}\mathcal{N}\left(\mathbf{m}_{nj}\vert\boldsymbol{\mu}_{jk},\boldsymbol{\Sigma}_{jk}\right),
\]
where $\boldsymbol{\Theta}\vcentcolon=\left\{ \pi_{jk},\boldsymbol{\mu}_{jk},\boldsymbol{\Sigma}_{jk}:\,j=1,\dots,M,k=1,\dots,K_{j}\right\} $
are the GMM parameters. Allowing GMM to represent the library, we
can get multiple Gaussian components for each endmember. Take the
dataset in Section~\ref{sec:Dataset} as an example, which can be
viewed from two perspectives. Fig.~\ref{fig:gmm_scatter} shows the
pixels from all the validation ROIs, library endmembers, and corresponding
Gaussian components when projected to 2 dimensions. The method for
estimating GMM parameters will be discussed later, however we can
see that the ellipses formed by these parameters surround validation
pixels at multiple positions on the edge of the pixel cloud. The pixels
can be viewed as picking points within ellipses and combining these
points linearly. Fig.~\ref{fig:gmm_wv_refl} shows the Gaussian components
from the wavelength-reflectance perspective, where the centers of
Gaussian components and their variation patterns are shown as curves.
Compared to MESMA, which evaluates every spectrum in the library,
GMM tries to combine every center of Gaussian components, but allows
the center to move according to its corresponding variation pattern.

Following the distribution assumption, if $\left\{ \mathbf{m}_{nj}:\,j=1,\dots,M\right\} $
are independent and the noise is also sampled from a Gaussian $p\left(\mathbf{n}_{n}\right)=\mathcal{N}\left(\mathbf{n}_{n}\vert\mathbf{0},\mathbf{D}\right)$,
then $\mathbf{y}_{n}=\sum_{j}\mathbf{m}_{nj}\alpha_{nj}+\mathbf{n}_{n}$
implies that the pixel follows a distribution
\[
p\left(\mathbf{y}_{n}\vert\boldsymbol{\alpha}_{n},\boldsymbol{\Theta},\mathbf{D}\right)=\sum_{\mathbf{k}\in\mathcal{K}}\pi_{\mathbf{k}}\mathcal{N}\left(\mathbf{y}_{n}\vert\boldsymbol{\mu}_{n\mathbf{k}},\boldsymbol{\Sigma}_{n\mathbf{k}}\right),
\]
where $\mathcal{K}\vcentcolon=\left\{ 1,\dots,K_{1}\right\} \times\left\{ 1,\dots,K_{2}\right\} \times\cdots\times\left\{ 1,\dots,K_{M}\right\} $
is the Cartesian product of the $M$ index sets, $\mathbf{k}=\left(k_{1},\dots,k_{M}\right)\in\mathcal{K}$,
$\pi_{\mathbf{k}}\in\mathbb{R}$, $\boldsymbol{\mu}_{n\mathbf{k}}\in\mathbb{R}^{B}$,
$\boldsymbol{\Sigma}_{n\mathbf{k}}\in\mathbb{R}^{B\times B}$ are
defined by 

\[
\pi_{\mathbf{k}}=\prod_{j=1}^{M}\pi_{jk_{j}},\,\boldsymbol{\mu}_{n\mathbf{k}}=\sum_{j=1}^{M}\alpha_{nj}\boldsymbol{\mu}_{jk_{j}},\,\boldsymbol{\Sigma}_{n\mathbf{k}}=\sum_{j=1}^{M}\alpha_{nj}^{2}\boldsymbol{\Sigma}_{jk_{j}}+\mathbf{D}.
\]

If we assume each pixel is independently sampled, the probability
density function of all the pixels is the product as

\[
p\left(\mathbf{Y}\vert\mathbf{A},\boldsymbol{\Theta},\mathbf{D}\right)=\prod_{n=1}^{N}p\left(\mathbf{y}_{n}\vert\boldsymbol{\alpha}_{n},\boldsymbol{\Theta},\mathbf{D}\right),
\]
where $\mathbf{A}\vcentcolon=\left[\boldsymbol{\alpha}_{1},\dots,\boldsymbol{\alpha}_{N}\right]^{T}\in\mathbb{R}^{N\times M}$.
Given $\mathbf{Y},\boldsymbol{\Theta},\mathbf{D}$, the abundances
$\mathbf{A}$ can be estimated by \emph{maximum likelihood estimation}
(MLE). Specifically, we want to maximize $p\left(\mathbf{Y}\vert\mathbf{A},\boldsymbol{\Theta},\mathbf{D}\right)$,
or minimize $-\log p\left(\mathbf{Y}\vert\mathbf{A},\boldsymbol{\Theta},\mathbf{D}\right)$,
which becomes the following optimization problem by combining the
above equations

\[
\mathcal{E}\left(\mathbf{A}\right)=-\sum_{n=1}^{N}\log\sum_{\mathbf{k}\in\mathcal{K}}\pi_{\mathbf{k}}\mathcal{N}\left(\mathbf{y}_{n}\vert\boldsymbol{\mu}_{n\mathbf{k}},\boldsymbol{\Sigma}_{n\mathbf{k}}\right),
\]

\[
\text{s.t.}\,\alpha_{nj}\ge0,\,\sum_{j=1}^{M}\alpha_{nj}=1,\,\forall n.
\]

The objective function can be minimized by a generalized \emph{expectation
maximization} (EM) algorithm, which alternates between an E step and
an M step \cite{meng1993maximum}. The E step calculates the posterior
probability of the latent variable given the observed data and old
parameters. The M step increases the expected value of the complete
data log-likelihood. In our case, the E step calculates

\begin{equation}
\gamma_{n\mathbf{k}}=\frac{\pi_{\mathbf{k}}\mathcal{N}\left(\mathbf{y}_{n}\vert\boldsymbol{\mu}_{n\mathbf{k}},\boldsymbol{\Sigma}_{n\mathbf{k}}\right)}{\sum_{\mathbf{k}\in\mathcal{K}}\pi_{\mathbf{k}}\mathcal{N}\left(\mathbf{y}_{n}\vert\boldsymbol{\mu}_{n\mathbf{k}},\boldsymbol{\Sigma}_{n\mathbf{k}}\right)}.\label{eq:E-step-gmm}
\end{equation}
The M step tries to minimize
\[
\mathcal{E}_{M}=-\sum_{n=1}^{N}\sum_{\mathbf{k}\in\mathcal{K}}\gamma_{n\mathbf{k}}\left\{ \log\pi_{\mathbf{k}}+\log\mathcal{N}\left(\mathbf{y}_{n}\vert\boldsymbol{\mu}_{n\mathbf{k}},\boldsymbol{\Sigma}_{n\mathbf{k}}\right)\right\} .
\]
It does not have a closed form solution for $\mathbf{A}$. But we
can use gradient descent to minimize $\mathcal{E}_{M}$, where the
derivative can be calculated by

\begin{equation}
\frac{\partial\mathcal{E}_{M}}{\partial\mathbf{A}}=-\sum_{\mathbf{k}\in\mathcal{K}}\boldsymbol{\Lambda}_{\mathbf{k}}\mathbf{R}_{\mathbf{k}}^{T}-2\mathbf{A}\circ\sum_{\mathbf{k}\in\mathcal{K}}\boldsymbol{\Psi}_{\mathbf{k}}\mathbf{S}_{\mathbf{k}}^{T},\label{eq:M-step_A1-gmm}
\end{equation}
where $\mathbf{R}_{\mathbf{k}}\in\mathbb{R}^{M\times B}$, $\mathbf{S}_{\mathbf{k}}\in\mathbb{R}^{M\times B^{2}}$
are defined by 
\[
\mathbf{R}_{\mathbf{k}}=\left[\boldsymbol{\mu}_{1k_{1}},\boldsymbol{\mu}_{2k_{2}},\dots,\boldsymbol{\mu}_{Mk_{M}}\right]^{T},
\]
\[
\mathbf{S}_{\mathbf{k}}=\left[\text{vec}\left(\boldsymbol{\Sigma}_{1k_{1}}\right),\text{vec}\left(\boldsymbol{\Sigma}_{2k_{2}}\right),\dots,\text{vec}\left(\boldsymbol{\Sigma}_{Mk_{M}}\right)\right]^{T},
\]
and $\boldsymbol{\Lambda}_{\mathbf{k}}\in\mathbb{R}^{N\times B}$,
$\boldsymbol{\Psi}_{\mathbf{k}}\in\mathbb{R}^{N\times B^{2}}$ denote
\[
\boldsymbol{\Lambda}_{\mathbf{k}}=\left[\boldsymbol{\lambda}_{1\mathbf{k}},\boldsymbol{\lambda}_{2\mathbf{k}},\dots,\boldsymbol{\lambda}_{N\mathbf{k}}\right]^{T},
\]
\[
\boldsymbol{\Psi}_{\mathbf{k}}=\left[\text{vec}\left(\boldsymbol{\Psi}_{1\mathbf{k}}\right),\text{vec}\left(\boldsymbol{\Psi}_{2\mathbf{k}}\right),\dots,\text{vec}\left(\boldsymbol{\Psi}_{N\mathbf{k}}\right)\right]^{T},
\]
where $\boldsymbol{\lambda}_{n\mathbf{k}}\in\mathbb{R}^{B\times1}$
and $\boldsymbol{\Psi}_{n\mathbf{k}}\in\mathbb{R}^{B\times B}$ are 

\[
\boldsymbol{\lambda}_{n\mathbf{k}}=\gamma_{n\mathbf{k}}\boldsymbol{\Sigma}_{n\mathbf{k}}^{-1}\left(\mathbf{y}_{n}-\boldsymbol{\mu}_{n\mathbf{k}}\right),
\]
\[
\boldsymbol{\Psi}_{n\mathbf{k}}=\frac{1}{2}\gamma_{n\mathbf{k}}\boldsymbol{\Sigma}_{n\mathbf{k}}^{-T}\left(\mathbf{y}_{n}-\boldsymbol{\mu}_{n\mathbf{k}}\right)\left(\mathbf{y}_{n}-\boldsymbol{\mu}_{n\mathbf{k}}\right)^{T}\boldsymbol{\Sigma}_{n\mathbf{k}}^{-T}-\frac{1}{2}\gamma_{n\mathbf{k}}\boldsymbol{\Sigma}_{n\mathbf{k}}^{-T}.
\]
Given an initial $\mathbf{A}$, we can update $\gamma_{n\mathbf{k}}$
and $\mathbf{A}$ alternately until convergence, which leads to a
local minimum of the objective function. This algorithm requires several
clarifications and we will explain them in the following subsections.

\subsection{Determining the GMM Parameters\label{subsec:Determine-the-GMM}}

Suppose we have a library of endmember spectra $\left\{ \mathbf{Y}_{j}\in\mathbb{R}^{N_{j}\times B}:\,j=1,\dots,M\right\} $,
with which we can estimate the GMM parameters $\boldsymbol{\Theta}$.
The difficulty comes from estimating the number of components $K_{j}$
for each endmember, as once we know $K_{j}$, $\left\{ \pi_{jk},\boldsymbol{\mu}_{jk},\boldsymbol{\Sigma}_{jk}\right\} $
can be estimated by the standard EM algorithm. Estimating this $K_{j}$
is sometimes called \emph{model selection} and has several approaches
\cite{mclachlan2014number}. We will use cross-validation-based information
criterion (CVIC) \cite{smyth2000model} as our metric to select $K_{j}$.

Given a candidate $K_{j}$, we can evaluate CVIC in the following
way. Let $\mathbf{Y}_{j}$ be the spectra for the $j$th endmember
in the library, we can divide them into $V=5$ subsets with equal
size. For each subset $\mathbf{Y}_{j}^{v}$, the remaining spectra
are input to a MLE with $K_{j}$ Gaussian components, and the trained
parameters are used to evaluate the log-likelihood of $\mathbf{Y}_{j}^{v}$.
Then the sum of all these log-likelihood values is calculated as $\mathcal{L}_{K_{j}}$,
which is our CVIC. Finally, the optimal $K_{j}$ is the one that maximizes
$\mathcal{L}_{K_{j}}$ out of all the candidates. To avoid many components,
we tried $K_{j}=1,2,3,4$.

This approach can serve as an ideal model selection. However, the
number of combinations $\left|\mathcal{K}\right|=\prod_{j}K_{j}$
can still be very large, especially in real datasets where the libraries
contain many spectra. Hence, we use a threshold $T_{CVIC}$ to further
reduce $K_{j}$ manually. Let $\mathcal{L}_{j}^{\prime}$ be the maximum
CVIC for the $j$th endmember; we pick the smallest $K_{j}$ such
that $\left|\mathcal{L}_{K_{j}}-\mathcal{L}_{j}^{\prime}\right|\le T_{CVIC}\mathcal{L}_{j}^{\prime}$.
Hence when $T_{CVIC}=0$, we have the ideal CVIC-based model selection.
As $T_{CVIC}$ increases, we can have a reduced number of components.

\subsection{Projection}

Analyzing the computation in Section~\ref{subsec:The-Gaussian-mixture},
we see that the time complexity is $O\left(\left|\mathcal{K}\right|NB^{3}\right)$
\cite{zhou2018gmmJournal}. In addition to the number of combinations,
the number of bands is also crucial to execution time. We can reduce
the time cost by reducing the dimensionality of the data.

We use PCA to reduce the dimensionality, which gives a center $\mathbf{c}\in\mathbb{R}^{B}$
and a projection matrix $\mathbf{E}\in\mathbb{R}^{B\times d}$ such
that all the spectra are processed as $\mathbf{E}^{T}\left(\mathbf{y}-\mathbf{c}\right)$.
Note that \eqref{eq:LMM_n} still holds if both the pixel spectra
and endmember spectra are projected in this way. Hence the estimated
abundances for the projected spectra are the original abundances for
the data. Also note that if an endmember follows a GMM distribution,
the projected endmember also follows a GMM distribution. So we can
directly estimate the GMM parameters from the projected library spectra.

As for the data input to find this projection, there are two possibilities.
One is to use all the pixel data. This works if the image is big enough,
such that all the endmembers have sufficient presence. However, if
the image contains fewer pixels (e.g. in the 16 m dataset) with limited
endmembers, the directions determined by PCA will present the variation
within the image, which may not distinguish different endmembers.
The other method is to use the spectral library, i.e. combine all
the spectra in the library and perform PCA on them. We adopted this
method for our dataset. Specifically, we selected an equal number
of spectra for each endmember class in the library and concatenated
them. This ensures that the relative sizes of endmember classes do
not affect the direction, and also ensures that the mean lies in the
center.

\subsection{The Algorithm}

The implementation of the algorithm is described in Algorithm~\ref{Algo-GMM}.
In step 1, the spectra in the library are concatenated to form an
input to PCA. We project the data to 10 dimensions in step 2. Step
3 is elaborated in Section~\ref{subsec:Determine-the-GMM}. Step
4 involves initialization of $\mathbf{A}$, which utilizes the information
of multiple means from the Gaussian components. To be specific, we
set $\boldsymbol{\alpha}_{n\mathbf{k}}\leftarrow\left(\mathbf{R}_{\mathbf{k}}\mathbf{R}_{\mathbf{k}}^{T}+\epsilon\mathbf{I}_{M}\right)^{-1}\mathbf{R}_{\mathbf{k}}\mathbf{y}_{n}$,
project $\boldsymbol{\alpha}_{n\mathbf{k}}$ onto the simplex space,
and initialize $\boldsymbol{\alpha}_{n}\leftarrow\boldsymbol{\alpha}_{n\mathbf{\hat{k}}}$
where $\hat{\mathbf{k}}=\arg\min_{\mathbf{k}}\Vert\mathbf{y}_{n}-\mathbf{R}_{\mathbf{k}}^{T}\boldsymbol{\alpha}_{n\mathbf{k}}\Vert^{2}$.
Step 5 is the main body, in which the M step is the most complicated
part. Because of the constraints on $\boldsymbol{\alpha}_{n}$, we
use projected gradient descent here. The projection function can be
found in \cite{duchi2008efficient,zhou2016spatial}. The step size
$\tau$ can be set adaptively by using a small initial value and gradually
increasing it by a multiplier of 10 as long as the objective function
keeps decreasing. Since the covariance matrix from endmember variability
is usually much larger than the noise covariance, the latter can be
negligible in the computation and we use $\mathbf{D}=0.001^{2}\mathbf{I}_{B}$.

\begin{algorithm}
\caption{Spectral unmixing with GMM}

Input: $\mathbf{Y}=\left[\mathbf{y}_{1},...,\mathbf{y}_{N}\right]^{T}$,
$\left\{ \mathbf{L}_{j}:\,j=1,\dots,M\right\} $, $T_{CVIC}$.
\begin{enumerate}
\item Determine the projection matrix by PCA.
\item Project the pixels and the spectra in the library to a low dimensional
subspace.
\item Estimate the numbers of components $\left\{ K_{j}\right\} $ using
CVIC and estimate the GMM parameters $\boldsymbol{\Theta}$ using
standard EM.
\item Initialize $\mathbf{A}$ by choosing the $\boldsymbol{\alpha}_{n\mathbf{k}}$
that minimizes the reconstruction error.
\item Alternately update the E step and M step until convergence.

\begin{itemize}
\item E step: update $\gamma_{n\mathbf{k}}$ by \eqref{eq:E-step-gmm}.
\item M step: update $\mathbf{A}$ by $\phi\left(\mathbf{A}-\tau\frac{\partial\mathcal{E}_{M}}{\partial\mathbf{A}}\right)$
where $\frac{\partial\mathcal{E}_{M}}{\partial\mathbf{A}}$ is defined
in \eqref{eq:M-step_A1-gmm}, $\tau$ is some step size, $\phi$ is
the projection function to the simplex space.
\end{itemize}
\end{enumerate}
Output: $\mathbf{A}$.

\label{Algo-GMM}
\end{algorithm}

\section{Results}

\subsection{Setup}

We ran GMM on the 16 m and 4 m images after training it using the
same resolution spectral libraries. Since GMM takes spectral libraries
to estimate the Gaussian components, we used the same components on
all 64 images. For reproducibility, we ran model selection of GMM
15 times, selected the most frequent combination, and applied it to
the dataset.

For comparison, we ran 2 set-based methods and 4 distribution-based
methods:
\begin{enumerate}
\item MESMA \cite{roberts1998mapping}. It was implemented in IDL by the
original authors and provided as an extension Viper Tool to ENVI.
We used the same parameters as in \cite{wetherley2017mapping}, i.e.
maximum RMSE 2.5\%, threshold RMSE 0.7\%, abundances constrained between
0 and 1, maximum shade threshold 80\%, and a maximum of three endmembers
plus shade for each pixel. Also, it will not allow multiple spectra
from one endmember class in the mixture. The obtained fractions were
normalized to give the final abundances.
\item Alternate angle minimization (AAM) \cite{heylen2016hyperspectral}.
Its code was implemented in Matlab and downloaded from Rob Heylen\textquoteright s
website. It tries every subset of endmembers, iteratively updates
the spectrum index of each endmember such that the reconstruction
error is minimized given the rest selected spectra, and hence finds
the best combination and abundances. Since it uses projection to find
the combination, theoretically it should work faster. It is different
from MESMA in several ways. First, it may not find the global minimum
because of its alternate optimization. Second, it may find a pixel
mixed by many endmembers instead of maximum three. Finally, it does
not include a shade endmember to adjust for brightness differences
between endmember and measured spectra. 
\item Gaussian mixture model (GMM). We used $T_{CVIC}=0.05$ for the 16
m dataset, which produced 216 combinations from the library. Because
the 4 m dataset was about 20 times larger, and the number of spectra
in the library was 3 times larger, we used a larger $T_{CVIC}=0.2$
(18 combinations) such that the whole process could still run in a
few hours.
\item GMM-1. We set the number of combinations to be 1, i.e. one component
for each endmember, which makes it to be NCM theoretically. However,
it has the same implementation of GMM hence reflects the difference
introduced by bringing multiple components.
\item Normal compositional model with sampling optimization (NCM Sampling).
There are many variations of NCM, with different optimization approaches
\cite{eches2010bayesian,eches2010estimating,halimi2015unsupervised,stein2003application,zare2010pce,zhangpso}.
We chose the sampling strategy in \cite{zare2013sampling} which does
not assume statistical independence between different bands.
\item Beta compositional model (BCM) \cite{du2014spatial}. It is available
from Alina Zare's website. Assuming the independence of bands, it
uses Beta distribution to model each band and unmixes the pixels.
\end{enumerate}
Excluding MESMA, which was implemented in IDL, all methods were implemented
in Matlab. MESMA was run on a PC with Intel Core i7-2760QM CPU and
8 GB memory. The other methods were run on a PC with Intel Core i7-3820
CPU and 64 GB memory. For distribution-based methods, the original
libraries were input to train the parameters for unmixing while set-based
methods used the reduced libraries.

We used two metrics to measure the differences between the estimated
and ground truth fractions: mean absolute difference (MAD) and correlation
coefficient (\emph{R}). They were calculated for each endmember class
based on the 64 pairs of values. To visualize the values, we used
a variation of the Bland-Altman plot where the x-axis is the ground
truth value and the y-axis is the estimated minus the ground truth
value \cite{bland1986statistical}. When comparing different algorithms
for unmixing quality, we will mainly resort to MAD as correlation
coefficient itself is not sufficient (slope and intercept are needed
to accompany \emph{R}).

\subsection{Accuracy and Efficiency\label{subsec:Accuracy-and-Efficiency}}

\textbf{16 m Case}. Table~\ref{table:accuracy_16m} shows the MAD
and correlation coefficient for the 16 m images. Original errors for
6 classes implies that GMM and AAM have the best accuracy, followed
by MESMA. The difference comes from the paved, roof and tree classes,
where GMM outperformed MESMA. In general, MESMA, AAM, GMM and GMM-1
had similar accuracy. Comparing all the distribution-based methods
with same input, GMM has the best performance overall, with fewest
errors for NPV, paved and roof.

Merged errors show that GMM and AAM retain their higher accuracy,
with MESMA falling further behind due to poor impervious accuracy.
Since merged errors are differences between summed up quantities of
similar materials, such as paved and roof, if the errors are enlarged
compared to individual errors, it means that both of the similar materials
are overestimated or underestimated, i.e. the algorithm tends for
confuse even dissimilar materials, such as pervious and impervious.

\begin{table*}
\centering

\caption{Comparison of error and correlation coefficient for the 16 m images\tnote{b}}

\begin{threeparttable}\footnotesize
\begin{centering}
\begin{tabular}{|c|c|c|c|c|c|c|c|}
\hline 
 &
 &
\multicolumn{2}{c|}{Set-based} &
\multicolumn{4}{c|}{Distribution-based}\tabularnewline
\hline 
\noalign{\vskip0.05cm}
\hline 
 &
MAD / $R^{2}$ &
MESMA &
AAM &
GMM &
GMM-1 &
NCM Sampling &
BCM\tabularnewline
\hline 
\noalign{\vskip0.05cm}
\hline 
\multirow{7}{*}{\begin{turn}{90}
Individual
\end{turn}} &
Turfgrass &
\textcolor{red}{0.029 / }\textcolor{red}{\emph{0.693}} &
\textcolor{red}{0.029 / }\textcolor{red}{\emph{0.703}} &
0.045 / \emph{0.632} &
0.041 / \emph{0.629} &
0.193 / \emph{0.117} &
0.042 / \emph{0.610}\tabularnewline
\cline{2-8} 
 & NPV &
\textcolor{red}{0.069 / }\textcolor{red}{\emph{0.830}} &
0.069 / \emph{0.819} &
\textcolor{red}{0.064 / }\textcolor{red}{\emph{0.805}} &
0.071 / \emph{0.813} &
0.132 / \emph{0.381} &
0.073 / \emph{0.750}\tabularnewline
\cline{2-8} 
 & Paved &
0.093 / \emph{0.588} &
\textcolor{red}{0.093 / }\textcolor{red}{\emph{0.685}} &
\textcolor{red}{0.081 / }\textcolor{red}{\emph{0.691}} &
0.093 / \emph{0.768} &
0.196 / \emph{0.058} &
0.096 / \emph{0.538}\tabularnewline
\cline{2-8} 
 & Roof &
0.087 / \emph{0.240} &
\textcolor{red}{0.079 / }\textcolor{red}{\emph{0.241}} &
\textcolor{red}{0.078 / }\textcolor{red}{\emph{0.279}} &
0.105 / \emph{0.032} &
0.112 / \emph{0.362} &
0.125 / \emph{0.000}\tabularnewline
\cline{2-8} 
 & Soil &
0.069 / \emph{0.773} &
0.080 / \emph{0.768} &
0.071 / \emph{0.834} &
\textcolor{red}{0.067 / }\textcolor{red}{\emph{0.781}} &
0.071 / \emph{0.000} &
\textcolor{red}{0.067 / }\textcolor{red}{\emph{0.733}}\tabularnewline
\cline{2-8} 
 & Tree &
0.098 / \emph{0.835} &
\textcolor{red}{0.065 / }\textcolor{red}{\emph{0.855}} &
0.076 / \emph{0.798} &
\textcolor{red}{0.071 / }\textcolor{red}{\emph{0.849}} &
0.107 / \emph{0.632} &
0.184 / \emph{0.534}\tabularnewline
\cline{2-8} 
 & \textbf{Average} &
\textbf{0.074} / \textbf{\emph{0.660}} &
\textbf{\textcolor{red}{0.069}}\textcolor{red}{{} / }\textbf{\textcolor{red}{\emph{0.678}}}\tnote{a} &
\textbf{\textcolor{red}{0.069}}\textcolor{red}{{} / }\textbf{\textcolor{red}{\emph{0.673}}} &
\textbf{0.075} / \textbf{\emph{0.645}} &
\textbf{0.135} / \textbf{\emph{0.258}} &
\textbf{0.098} / \textbf{\emph{0.527}}\tabularnewline
\hline 
\noalign{\vskip0.05cm}
\hline 
\multirow{4}{*}{\begin{turn}{90}
Merged
\end{turn}} &
GV &
0.088 / \emph{0.909} &
\textcolor{red}{0.057 / }\textcolor{red}{\emph{0.898}} &
\textcolor{red}{0.070 / }\textcolor{red}{\emph{0.872}} &
0.073 / \emph{0.913} &
0.156 / \emph{0.515} &
0.175 / \emph{0.701}\tabularnewline
\cline{2-8} 
 & Pervious &
0.082 / \emph{0.836} &
0.084 / \emph{0.870} &
\textcolor{red}{0.072 / }\textcolor{red}{\emph{0.880}} &
0.075 / \emph{0.880} &
0.145 / \emph{0.515} &
\textcolor{red}{0.074 / }\textcolor{red}{\emph{0.886}}\tabularnewline
\cline{2-8} 
 & Impervious &
0.102 / \emph{0.762} &
\textcolor{red}{0.087 / }\textcolor{red}{\emph{0.818}} &
\textcolor{red}{0.084 / }\textcolor{red}{\emph{0.845}} &
0.088 / \emph{0.819} &
0.199 / \emph{0.181} &
0.175 / \emph{0.722}\tabularnewline
\cline{2-8} 
 & \textbf{Average} &
\textbf{0.091} / \textbf{\emph{0.836}} &
\textbf{\textcolor{red}{0.076}}\textcolor{red}{{} / }\textbf{\textcolor{red}{\emph{0.862}}} &
\textbf{\textcolor{red}{0.075}}\textcolor{red}{{} / }\textbf{\textcolor{red}{\emph{0.866}}} &
\textbf{\textcolor{black}{0.079}}\textcolor{black}{{} / }\textbf{\textcolor{black}{\emph{0.871}}} &
\textbf{0.167} / \textbf{\emph{0.404}} &
\textbf{0.141} / \textbf{\emph{0.770}}\tabularnewline
\hline 
\end{tabular}
\par\end{centering}
\begin{tablenotes} \item [a] the entries in red fonts denote the best two results in each category. \item [b] the running time on all the images for the 6 methods is 18, 27, 83, 0.3, 220, 101 minutes respectively.  \end{tablenotes} \end{threeparttable}

\label{table:accuracy_16m}
\end{table*}

\begin{figure*}
\begin{centering}
\includegraphics[width=16.5cm]{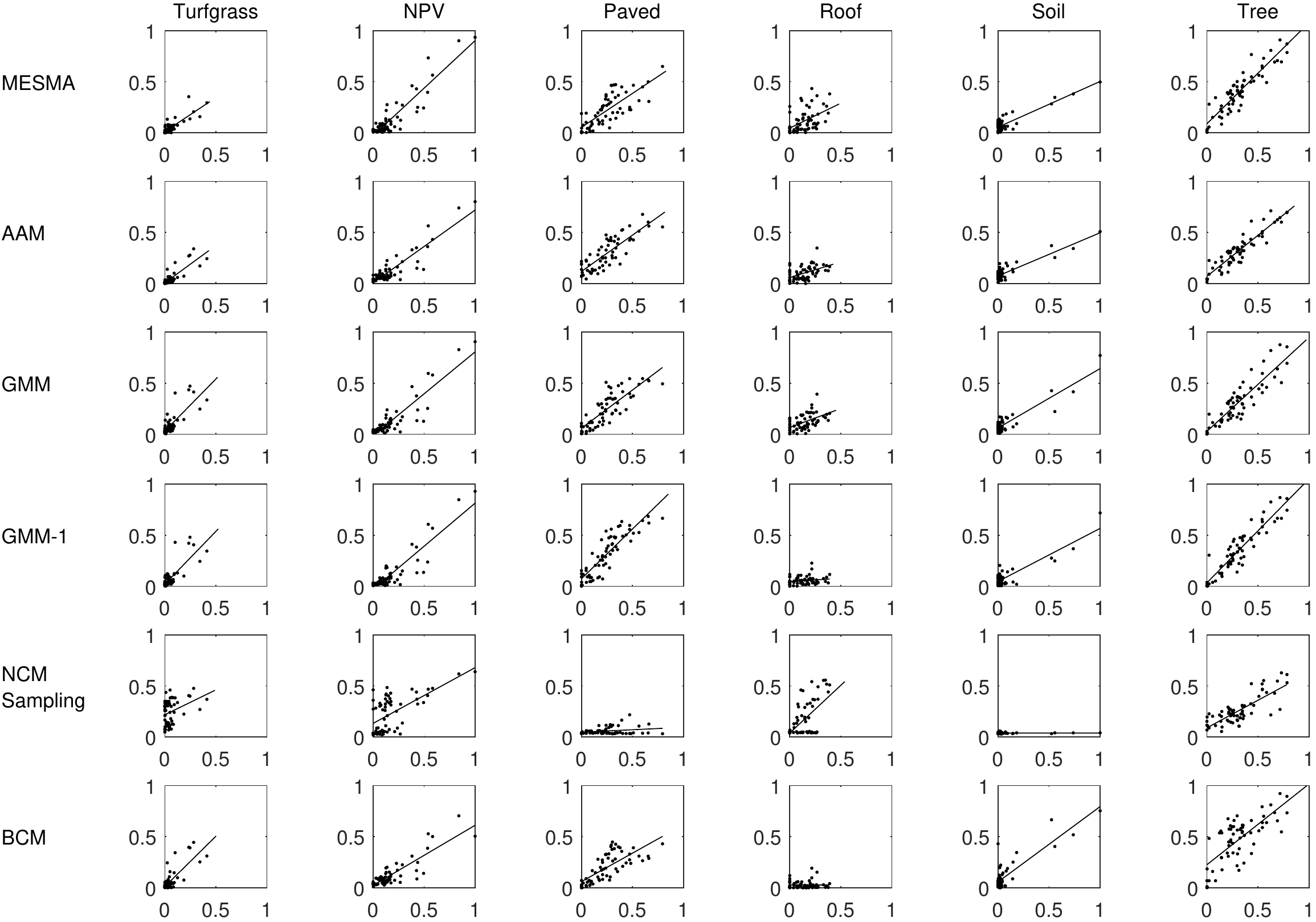}
\par\end{centering}
\caption{Scatter plots of 64 abundance values in 16 m for ground truth (x-axis)
and estimated (y-axis).}

\label{fig:abund_scatter_16m}
\end{figure*}

\begin{figure*}
\begin{centering}
\includegraphics[width=16.5cm]{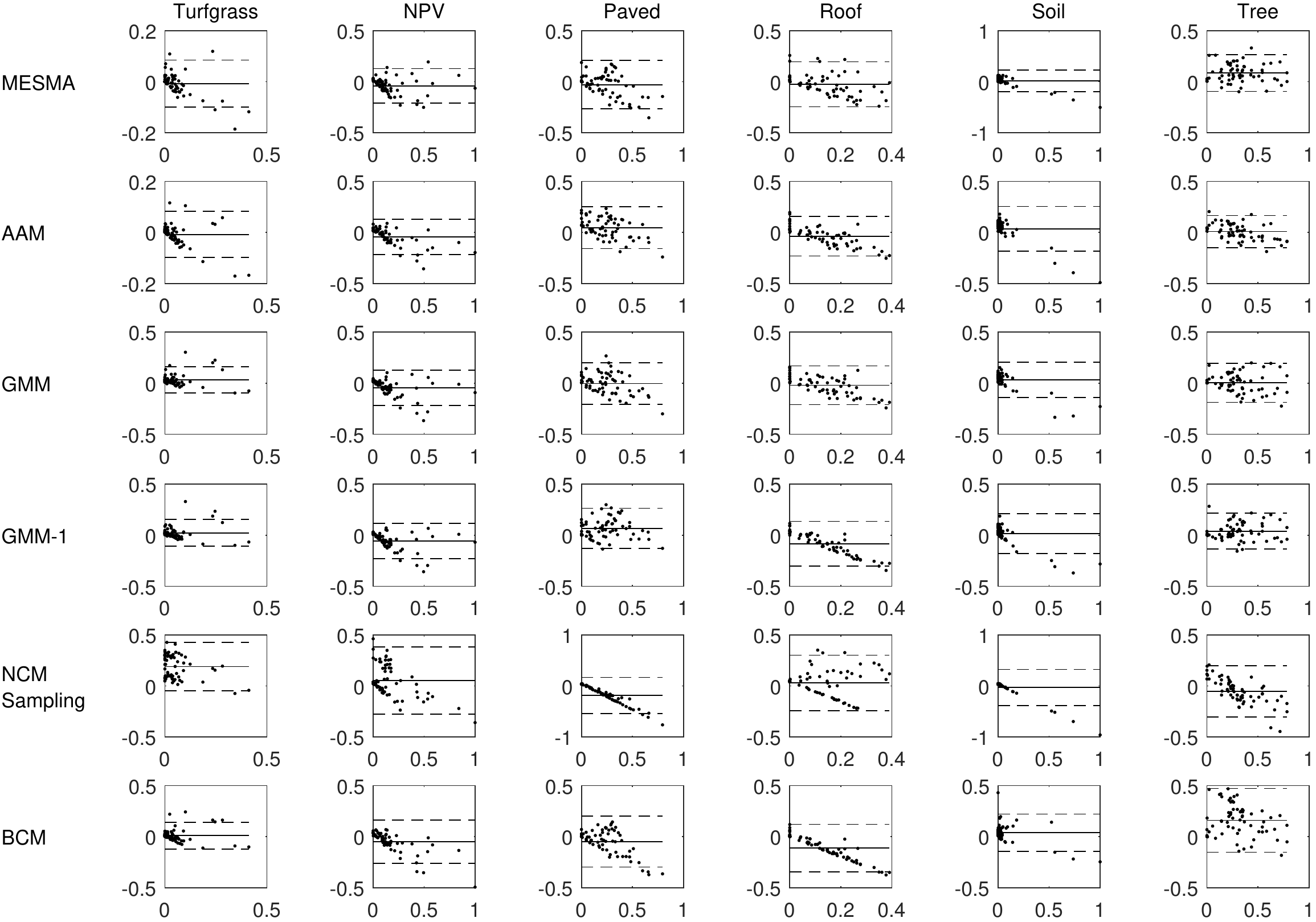}
\par\end{centering}
\caption[Bland-Altman plots of 64 abundance values in 16 m for ground truth
and estimated.]{Bland-Altman plots of 64 abundance values in 16 m for ground truth
and estimated. The x-axis is the ground truth. The y-axis is the difference
between estimated and ground truth. The solid line in each plot is
the mean of these differences while the dashed lines show the mean
plus (minus) twice the standard deviation of these differences.}

\label{fig:abund_bland_altman_16m}
\end{figure*}

Fig.~\ref{fig:abund_scatter_16m} compares the estimated total abundances
to validated abundances for each material. We can see that NCM Sampling
and BCM tend to ignore paved or roof when they have presence. The
difference statistics between the estimated and the ground truth are
plotted in the Bland-Altman plots in Fig.~\ref{fig:abund_bland_altman_16m},
where the set-based methods and GMM appear to be better than the others.

\textbf{4 m Case}. The error statistics of 4 m data are shown in Table~\ref{table:accuracy_4m}.
MESMA and GMM are the most accurate, with AAM is not as good as MESMA.
One possible reason for the high accuracy of MESMA is that MESMA inherently
takes shade into account while AAM only combines input library spectra.
Hence, the slightly better performance of GMM over AAM is more significant
since they both ignore shade.

\begin{table*}
\centering

\caption{Comparison of error and correlation coefficient for the 4 m images\tnote{b}}

\begin{threeparttable}\footnotesize
\begin{centering}
\begin{tabular}{|c|c|c|c|c|c|c|c|}
\hline 
 &
 &
\multicolumn{2}{c|}{Set-based} &
\multicolumn{4}{c|}{Distribution-based}\tabularnewline
\hline 
\noalign{\vskip0.05cm}
\hline 
 &
MAD / $R^{2}$ &
MESMA &
AAM &
GMM &
GMM-1 &
NCM Sampling &
BCM\tabularnewline
\hline 
\noalign{\vskip0.05cm}
\hline 
\multirow{7}{*}{\begin{turn}{90}
Individual
\end{turn}} &
Turfgrass &
\textcolor{red}{0.028 / }\textcolor{red}{\emph{0.740}} &
0.033 / \emph{0.712} &
0.036 / \emph{0.811} &
\textcolor{red}{0.027 / }\textcolor{red}{\emph{0.811}} &
0.033 / \emph{0.754} &
0.049 / \emph{0.710}\tabularnewline
\cline{2-8} 
 & NPV &
\textcolor{red}{0.044 / }\textcolor{red}{\emph{0.885}} &
0.059 / \emph{0.874} &
\textcolor{red}{0.052 / }\textcolor{red}{\emph{0.879}} &
0.052 / \emph{0.867} &
0.068 / \emph{0.872} &
0.074 / \emph{0.822}\tabularnewline
\cline{2-8} 
 & Paved &
\textcolor{red}{0.057 / }\textcolor{red}{\emph{0.859}} &
0.100 / \emph{0.822} &
\textcolor{red}{0.061 / }\textcolor{red}{\emph{0.834}} &
0.075 / \emph{0.794} &
0.165 / \emph{0.847} &
0.089 / \emph{0.607}\tabularnewline
\cline{2-8} 
 & Roof &
\textcolor{red}{0.060 / }\textcolor{red}{\emph{0.574}} &
0.078 / \emph{0.287} &
\textcolor{red}{0.069 / }\textcolor{red}{\emph{0.458}} &
0.100 / \emph{0.250} &
0.199 / \emph{0.655} &
0.105 / \emph{0.181}\tabularnewline
\cline{2-8} 
 & Soil &
\textcolor{red}{0.040 / }\textcolor{red}{\emph{0.941}} &
0.065 / \emph{0.793} &
0.078 / \emph{0.904} &
0.084 / \emph{0.897} &
\textcolor{red}{0.065 / }\textcolor{red}{\emph{0.867}} &
0.094 / \emph{0.734}\tabularnewline
\cline{2-8} 
 & Tree &
0.049 / \emph{0.910} &
0.055 / \emph{0.918} &
\textcolor{red}{0.043 / }\textcolor{red}{\emph{0.934}} &
\textcolor{red}{0.039 / }\textcolor{red}{\emph{0.937}} &
0.057 / \emph{0.926} &
0.102 / \emph{0.682}\tabularnewline
\cline{2-8} 
 & \textbf{Average} &
\textbf{\textcolor{red}{0.046}}\textcolor{red}{{} / }\textbf{\textcolor{red}{\emph{0.818}}}\tnote{a} &
\textbf{0.065} / \textbf{\emph{0.735}} &
\textbf{\textcolor{red}{0.056}}\textcolor{red}{{} / }\textbf{\textcolor{red}{\emph{0.803}}} &
\textbf{0.063} / \textbf{\emph{0.759}} &
\textbf{0.098} / \textbf{\emph{0.820}} &
\textbf{0.086} / \textbf{\emph{0.623}}\tabularnewline
\hline 
\noalign{\vskip0.05cm}
\hline 
\multirow{4}{*}{\begin{turn}{90}
Merged
\end{turn}} &
GV &
\textcolor{red}{0.039 / }\textcolor{red}{\emph{0.943}} &
0.045 / \emph{0.931} &
0.043 / \emph{0.943} &
\textcolor{red}{0.042 / }\textcolor{red}{\emph{0.942}} &
0.058 / \emph{0.932} &
0.120 / \emph{0.823}\tabularnewline
\cline{2-8} 
 & Pervious &
\textcolor{red}{0.053 / }\textcolor{red}{\emph{0.935}} &
0.091 / \emph{0.858} &
\textcolor{red}{0.075 / }\textcolor{red}{\emph{0.916}} &
0.082 / \emph{0.906} &
0.087 / \emph{0.886} &
0.078 / \emph{0.897}\tabularnewline
\cline{2-8} 
 & Impervious &
\textcolor{red}{0.054 / }\textcolor{red}{\emph{0.943}} &
0.072 / \emph{0.817} &
\textcolor{red}{0.067 / }\textcolor{red}{\emph{0.921}} &
0.075 / \emph{0.911} &
0.092 / \emph{0.895} &
0.151 / \emph{0.791}\tabularnewline
\cline{2-8} 
 & \textbf{Average} &
\textbf{\textcolor{red}{0.049}}\textcolor{red}{{} / }\textbf{\textcolor{red}{\emph{0.941}}} &
\textbf{0.069} / \textbf{\emph{0.868}} &
\textbf{\textcolor{red}{0.062 }}\textcolor{red}{/ }\textbf{\textcolor{red}{\emph{0.927}}} &
\textbf{\textcolor{black}{0.067}}\textcolor{black}{{} / }\textbf{\textcolor{black}{\emph{0.919}}} &
\textbf{0.079} / \textbf{\emph{0.904}} &
\textbf{0.116} / \textbf{\emph{0.837}}\tabularnewline
\hline 
\end{tabular}
\par\end{centering}
\begin{tablenotes} \item [a] the entries in red fonts denote the best two results in each category. \item [b] the running time on all the images for the 6 methods is 165, 499, 213, 8, 3153, 1347 minutes respectively.  \end{tablenotes} \end{threeparttable}

\label{table:accuracy_4m}
\end{table*}

\begin{figure*}
\begin{centering}
\includegraphics[width=16.5cm]{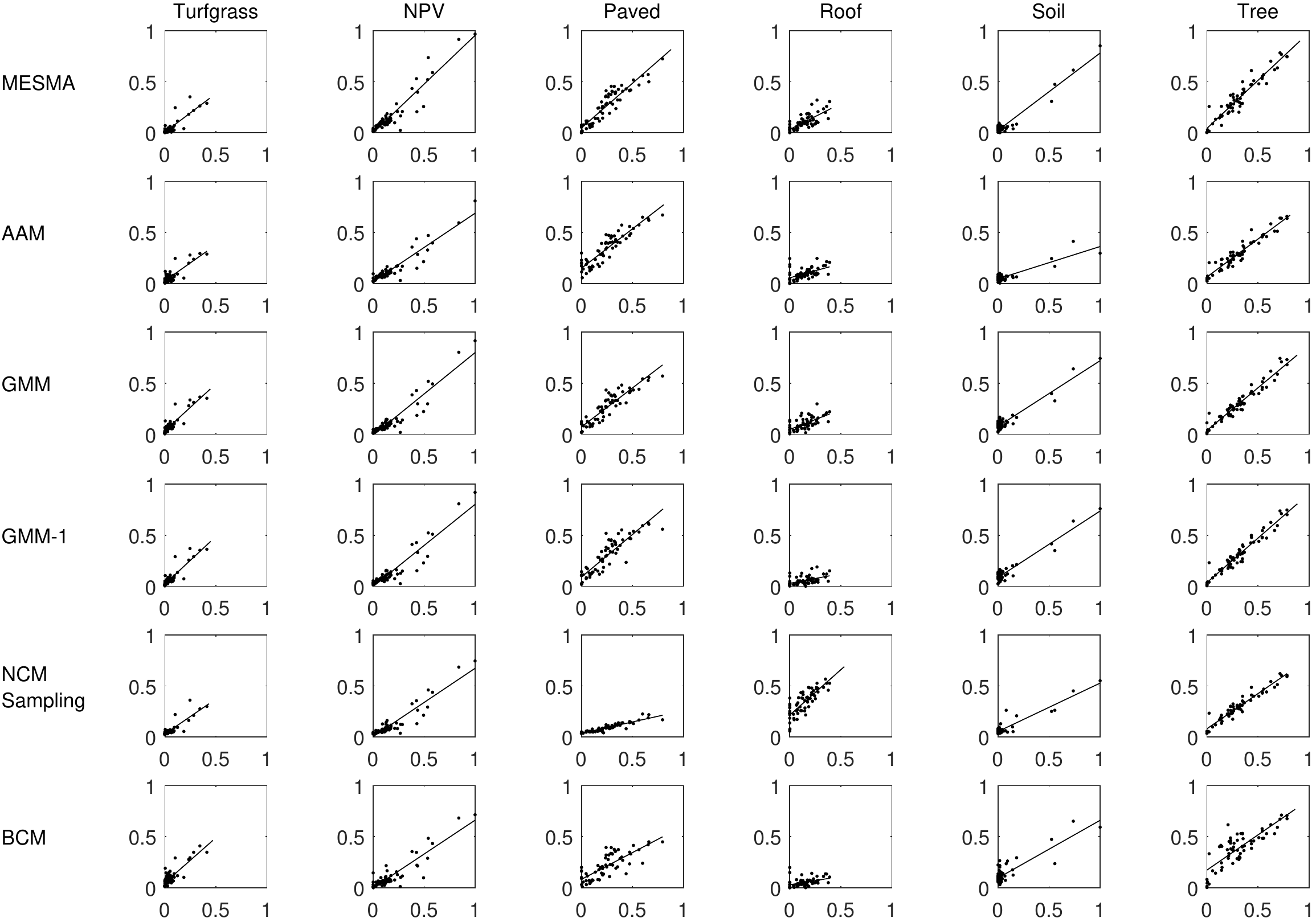}
\par\end{centering}
\caption{Scatter plots of 64 abundance values in 4 m for ground truth (x-axis)
and estimated (y-axis).}

\label{fig:abund_scatter_4m}
\end{figure*}

\begin{figure*}
\begin{centering}
\includegraphics[width=16.5cm]{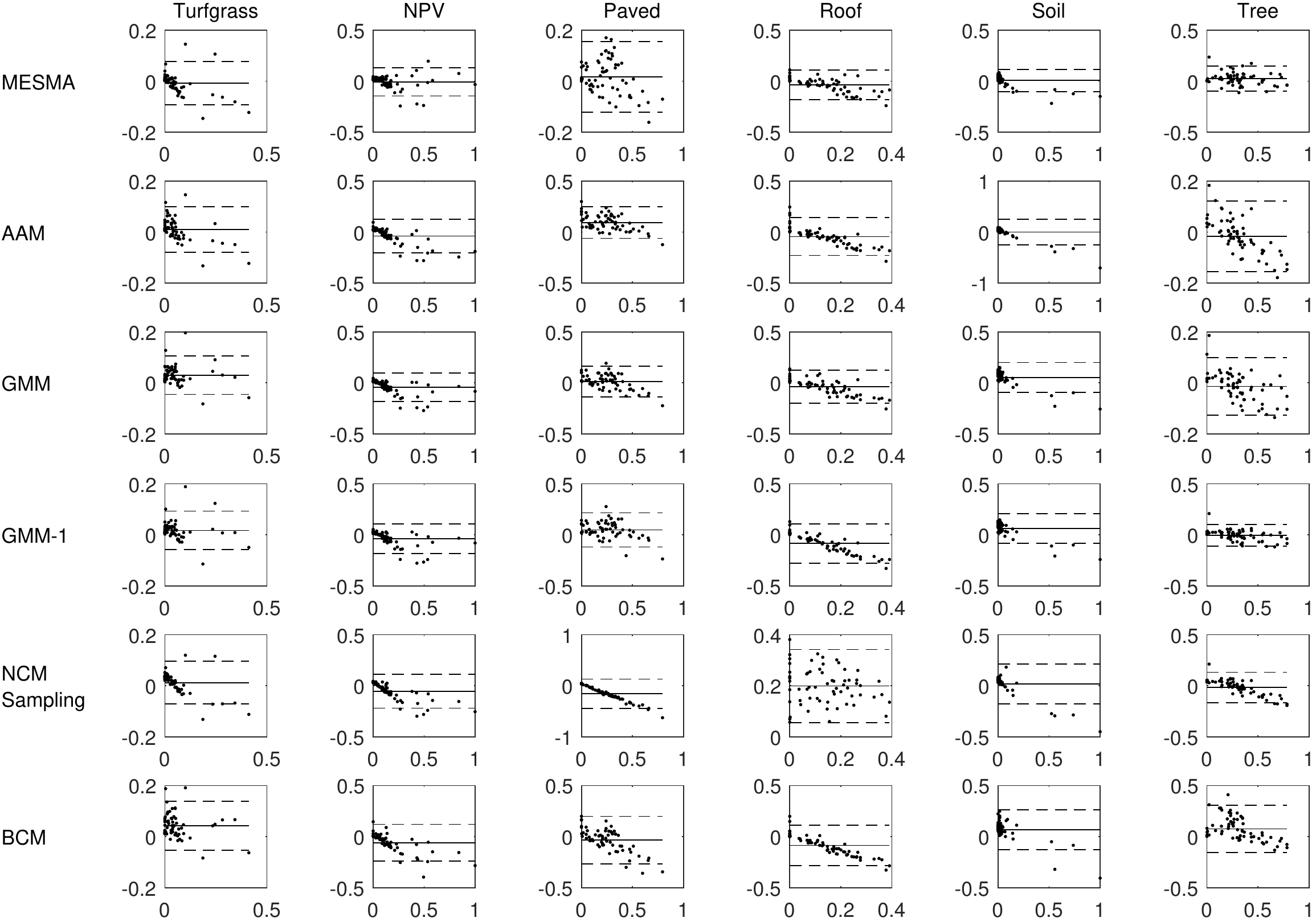}
\par\end{centering}
\caption{Bland-Altman plots of 64 abundance values in 4 m for ground truth
and estimated.}

\label{fig:abund_bland_altman_4m}
\end{figure*}

Fig.~\ref{fig:abund_scatter_4m} and Fig.~\ref{fig:abund_bland_altman_4m}
show the scatter plots and Bland-Altman plots for this data. Similar
to the statistics in Table~\ref{table:accuracy_4m}, GMM has an advantage
over the other distribution-based methods on paved and roof. Compared
to AAM, GMM presents stable results to some extremely bad outliers
for soil in AAM.

\textbf{Efficiency}. Since they were run on different machines with
different implementation (MESMA and NCM Sampling have multiple threads),
the time costs are not for comparison, but for reference. In general,
all the algorithms run in a few hours. The fastest algorithm is GMM-1,
which is our implemented GMM with only one combination. NCM Sampling
turns out to be the slowest algorithm. It is expected since sampling
algorithms are usually slower than deterministic algorithms. The times
costs on the 4 m dataset are usually more than 10 times slower than
those on the 16 m dataset. This is because the image size of the former
is 16-19 times larger than the latter and the library size is also
larger. The least gap comes from GMM because of a changed $T_{CVIC}$
leading to a significantly less number of combinations. In \cite{heylen2016hyperspectral}
the authors show that with the same implementation AAM is much faster
than MESMA, but here the result is converse. One reason is that the
parameters of MESMA force it to pick at most 3 endmembers for a pixel
instead of all the combinations. Also, multi-threading and implementation
techniques impact the real world time costs significantly.

\subsection{Extend to Semi-realistic Images}

We extended the experiments to semi-realistic images to check if the
algorithm implementation or library reduction was overfitted to this
particular dataset. The method was to test the algorithms on another
batch of synthetic images generated by the library spectra. Since
all the algorithms assumed that a pixel was a linear combination of
endmember spectra from the library, the creation of this synthetic
dataset would use this assumption.

\begin{figure*}
\begin{centering}
\includegraphics[width=18cm]{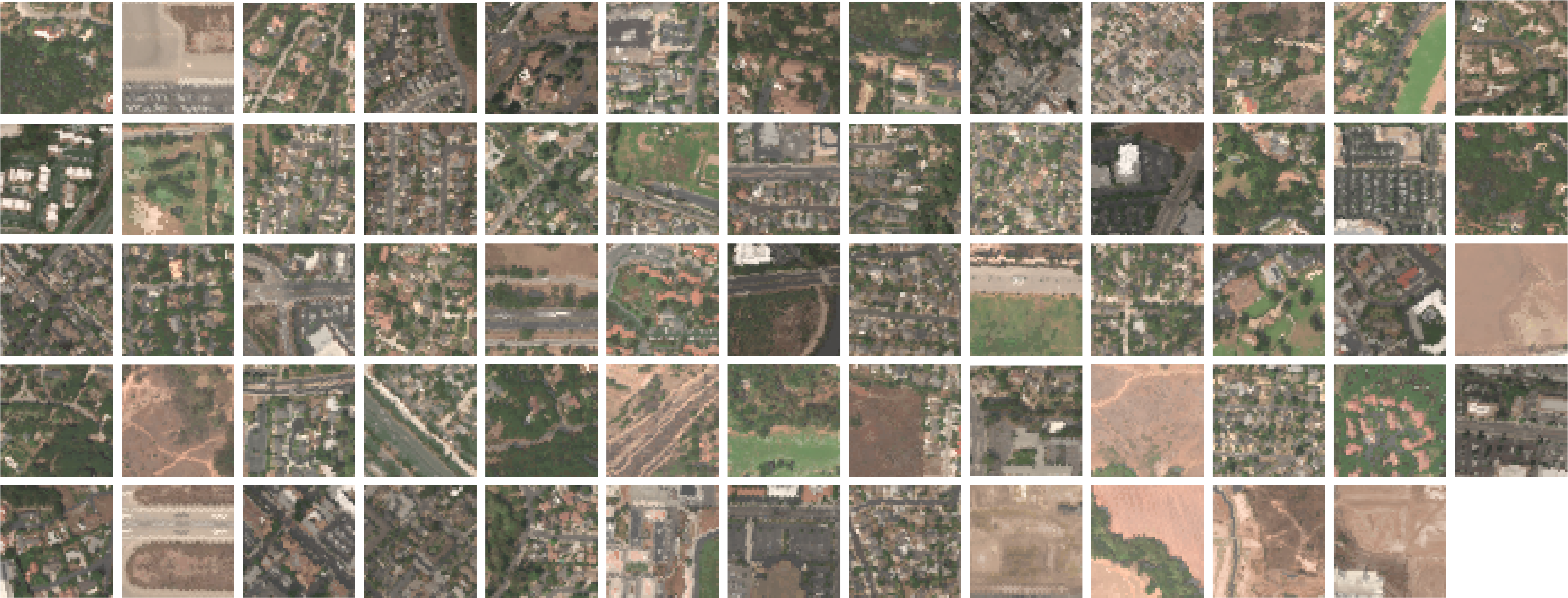}
\par\end{centering}
\caption[Simulated 4 m ROI images.]{Simulated 4 m ROI images. They are very similar to the real images
in Fig.~\ref{fig:validation_polygons}.}

\label{fig:synthetic_images}
\end{figure*}

We created this dataset following the literature that emphasizes realistic
simulation \cite{gao2013comparative,hao2015semi}. For each image
in the original dataset, we randomly sampled spectra from the full
library while the number of spectra for each endmember class is equal
to the number in the reduced library. Then we used AAM to unmix the
image with these sampled spectra. The obtained abundances were sorted
to keep the largest three while the other were set to 0, and rescaled
such that their summation was one. This is to conform with the assumption
of MESMA. The endmembers and abundances were combined according to
the LMM to generate pixels. In this way, we can have a dataset where
the endmembers are randomly picked from the full library, and the
spatial distribution of abundances looks similar to the original one.
Fig.~\ref{fig:synthetic_images} shows all the 4 m synthetic images
generated in this way. Comparing it with Fig.~\ref{fig:validation_polygons},
we can see its similarity. But inherently, the synthetic images follow
exactly the LMM with at most 3 endmembers for each pixel and they
are randomly picked from the full library.

We ran all the algorithms on this simulated dataset. Table~\ref{table:accuracy_syn}
shows the unmixing results on this dataset. We can see that GMM and
NCM Sampling turn out to be the best methods. The superior performance
of NCM contrasts sharply to its worst result in Table~\ref{table:accuracy_16m}.
Since we evaluate the difference between total abundances for a material,
it is possible that the relatively large pixel abundance error is
mitigated by averaging them. This is more possible for sampling algorithms
because statistically they tend to sample values around the correct
ones. 

\begin{table*}
\centering

\caption{Comparison of error and correlation coefficient for the synthetic
images\tnote{b}}

\begin{threeparttable}\footnotesize
\begin{centering}
\begin{tabular}{|c|c|c|c|c|c|c|c|}
\hline 
 &
 &
\multicolumn{2}{c|}{Set-based} &
\multicolumn{4}{c|}{Distribution-based}\tabularnewline
\hline 
\noalign{\vskip0.05cm}
\hline 
 &
MAD / $R^{2}$ &
MESMA &
AAM &
GMM &
GMM-1 &
NCM Sampling &
BCM\tabularnewline
\hline 
\noalign{\vskip0.05cm}
\hline 
\multirow{7}{*}{\begin{turn}{90}
16 m
\end{turn}} &
Turfgrass &
0.044 / \emph{0.791} &
0.050 / \emph{0.837} &
\textcolor{red}{0.021 / }\textcolor{red}{\emph{0.913}} &
0.025 / \emph{0.916} &
\textcolor{red}{0.016 / }\textcolor{red}{\emph{0.954}} &
0.042 / \emph{0.827}\tabularnewline
\cline{2-8} 
 & NPV &
0.039 / \emph{0.948} &
0.027 / \emph{0.940} &
\textcolor{red}{0.023 / }\textcolor{red}{\emph{0.963}} &
0.026 / \emph{0.956} &
\textcolor{red}{0.018 / }\textcolor{red}{\emph{0.978}} &
0.041 / \emph{0.848}\tabularnewline
\cline{2-8} 
 & Paved &
\textcolor{red}{0.074 / }\textcolor{red}{\emph{0.512}} &
0.115 / \emph{0.563} &
0.078 / \emph{0.745} &
0.116 /\emph{ 0.598} &
\textcolor{red}{0.038 / }\textcolor{red}{\emph{0.927}} &
0.101 / \emph{0.307}\tabularnewline
\cline{2-8} 
 & Roof &
0.063 / \emph{0.441} &
0.063 / \emph{0.569} &
\textcolor{red}{0.054 / }\textcolor{red}{\emph{0.742}} &
0.118 / \emph{0.259} &
\textcolor{red}{0.027 / }\textcolor{red}{\emph{0.894}} &
0.129 / \emph{0.170}\tabularnewline
\cline{2-8} 
 & Soil &
0.038 /\emph{ 0.643} &
0.033 / \emph{0.723} &
0.028 / \emph{0.738} &
\textcolor{red}{0.028 / }\textcolor{red}{\emph{0.781}} &
\textcolor{red}{0.016 / }\textcolor{red}{\emph{0.956}} &
0.067 / \emph{0.604}\tabularnewline
\cline{2-8} 
 & Tree &
0.090 / \emph{0.882} &
0.058 / \emph{0.822} &
\textcolor{red}{0.051 / }\textcolor{red}{\emph{0.880}} &
0.060 / \emph{0.923} &
\textcolor{red}{0.029 / }\textcolor{red}{\emph{0.960}} &
0.179 / \emph{0.669}\tabularnewline
\cline{2-8} 
 & \textbf{Average} &
\textbf{0.058} / \textbf{\emph{0.703}} &
\textbf{0.058} /\emph{ }\textbf{\emph{0.742}} &
\textbf{\textcolor{red}{0.042}}\textcolor{red}{{} / }\textbf{\textcolor{red}{\emph{0.830}}} &
\textbf{0.062} / \textbf{\emph{0.739}} &
\textbf{\textcolor{red}{0.024}}\textcolor{red}{{} / }\textbf{\textcolor{red}{\emph{0.945}}} &
\textbf{0.093} / \textbf{\emph{0.571}}\tabularnewline
\hline 
\noalign{\vskip0.05cm}
\hline 
\multirow{7}{*}{\begin{turn}{90}
4 m
\end{turn}} &
Turfgrass &
0.032 / \emph{0.838} &
0.019 / \emph{0.912} &
0.021 / \emph{0.921} &
\textcolor{red}{0.015 / }\textcolor{red}{\emph{0.921}} &
\textcolor{red}{0.014 / }\textcolor{red}{\emph{0.919}} &
0.019 / \emph{0.896}\tabularnewline
\cline{2-8} 
 & NPV &
0.040 /\emph{ 0.955} &
0.016 / \emph{0.965} &
\textcolor{red}{0.012 / }\textcolor{red}{\emph{0.990}} &
\textcolor{red}{0.013 / }\textcolor{red}{\emph{0.991}} &
\textcolor{black}{0.019 / }\textcolor{black}{\emph{0.995}} &
0.027 / \emph{0.955}\tabularnewline
\cline{2-8} 
 & Paved &
0.062 / \emph{0.836} &
0.121 / \emph{0.819} &
0.070 / \emph{0.773} &
0.103 / \emph{0.742} &
\textcolor{red}{0.041 / }\textcolor{red}{\emph{0.890}} &
\textcolor{red}{0.053 / }\textcolor{red}{\emph{0.723}}\tabularnewline
\cline{2-8} 
 & Roof &
0.048 / \emph{0.478} &
0.051 / \emph{0.537} &
\textcolor{red}{0.046 / }\textcolor{red}{\emph{0.472}} &
0.087 / \emph{0.254} &
\textcolor{red}{0.035 / }\textcolor{red}{\emph{0.617}} &
0.088 / \emph{0.222}\tabularnewline
\cline{2-8} 
 & Soil &
0.043 / \emph{0.661} &
0.042 / \emph{0.845} &
\textcolor{red}{0.026 / }\textcolor{red}{\emph{0.840}} &
\textcolor{red}{0.028 / }\textcolor{red}{\emph{0.813}} &
0.036 / \emph{0.953} &
0.041 / \emph{0.766}\tabularnewline
\cline{2-8} 
 & Tree &
0.050 / \emph{0.906} &
\textcolor{red}{0.035 / }\textcolor{red}{\emph{0.911}} &
0.055 / \emph{0.864} &
0.045 / \emph{0.880} &
\textcolor{red}{0.033 / }\textcolor{red}{\emph{0.951}} &
0.099 / \emph{0.748}\tabularnewline
\cline{2-8} 
 & \textbf{Average} &
\textbf{0.046} / \textbf{\emph{0.779}} &
\textbf{0.047} / \textbf{\emph{0.832}} &
\textbf{\textcolor{red}{0.038}}\textcolor{red}{{} / }\textbf{\textcolor{red}{\emph{0.810}}}\tnote{a} &
\textbf{0.049} / \textbf{\emph{0.767}} &
\textbf{\textcolor{red}{0.030}}\textcolor{red}{{} / }\textbf{\textcolor{red}{\emph{0.887}}} &
\textbf{0.054} / \textbf{\emph{0.718}}\tabularnewline
\hline 
\end{tabular}
\par\end{centering}
\begin{tablenotes} \item [a] the entries in red fonts denote the best two results in each category. \end{tablenotes} \end{threeparttable}

\label{table:accuracy_syn}
\end{table*}

Compared to the negligible difference between GMM and set-based methods
in Section~\ref{subsec:Accuracy-and-Efficiency}, the advantage of
GMM is more obvious. Since the endmembers were randomly sampled from
the big library, set-based methods were less capable to unmix the
pixels using a reduced library that was derived based on the another
dataset. It is possible that a different reduced library based on
this simulated dataset may lead to better results for set-based methods.
However, that means, set-based methods may not be as robust as GMM
across datasets.

\section{Discussion and Conclusion}

We have proposed an unmixing algorithm based on endmember variability
modeled by GMM distributions. We validated the algorithm on a dataset
consisting of 128 images across 2 scales, with ground truth abundances
obtained by inspecting high resolution images. The results show that
with large libraries, GMM achieves comparable accuracy to MESMA without
the need for guided manual library reduction. We will discuss several
issues regarding the dataset, algorithm and results in this Section.

The dataset was well developed with various scenes, but the ground
truth has intrinsic errors coming from UTM coordinates. It happens
when the universal coordinates are used in the hyperspectral images
and the other high-resolution images for region correspondence. Because
these airborne images are spatially calibrated from its unstable collection
process, the coordinates derived from the map information may not
accurately reflect the real coordinates. Therefore, the region for
calculating the total abundances may have a small shift compared to
the region in the hyperspectral image. This is more likely in the
16 m data, which may explain larger overall errors for all the algorithms,
and could be mitigated by registering the two images to find exact
correspondence in the future \cite{zhou2017nonrigidReg}.

In applying GMM on the dataset, there are several implementation details
that affected the results. First, the projection affected the results
of GMM. Hence, we used a carefully determined projection in Algorithm~\ref{Algo-GMM}.
Second, the number of combinations impacted the performance. If some
training data were present, we could gradually decrease $T_{CVIC}$
until the error stopped improving during reasonable time. Otherwise,
we suggest gradually increasing $T_{CVIC}$ from 0 (without running
the whole algorithm) until the number of combinations reduces to approximately
100, which is a reasonable time cost multiplier of GMM-1. Third, we
didn\textquoteright t use a spatial prior in the original GMM paper
\cite{zhou2018gmmJournal}. We found that the prior made it work better
on some images, while worse on some other images when applied to this
dataset. In total, it didn\textquoteright t improve the results much.
This has two possible reasons: (i) the pixel size is big enough such
that smoothness and sparsity are not obvious on the abundance maps;
(ii) the dataset contains a variety of scenes in which some of them
violate this property.

We also have some remarks on the results. For the real dataset, GMM
is slightly better than MESMA for the 16 m data while slightly inferior
for the 4 m data, so they are close in accuracy. However, MESMA used
a library reduction method that relied on manual guidance and user
knowledge of the study area, while GMM used the original library without
refinement. Hence, GMM may be more applicable to datasets without
ground truth. Second, the results of MESMA were slightly different
from those reported in \cite{wetherley2017mapping}, which used this
same imagery and validation dataset. In \cite{wetherley2017mapping},
the average $R^{2}$ for individual and merge cases are 0.642 and
0.867 for 16 m, 0.811 and 0.923 for 4 m. Comparing them to Table~\ref{table:accuracy_16m}
and Table~\ref{table:accuracy_4m}, they are very similar. Note that
we used Matlab to extract the polygon ROIs directly from the original
images while in \cite{wetherley2017mapping} the images were resampled
to a uniform spatial resolution of 18 m and corrected for locational
errors using Delaunay triangulation. Additionally, that study used
178 spectral bands in contrast to 164 bands used here.

Future work could include developing a dataset with more images and
more scale variation. Also, validation abundances can be more accurately
obtained by registration of hyperspectral and high resolution images.
Further work can also be done to improve the efficiency of GMM using
covariance matrices with a simple form.

\section*{Acknowledgments}

The authors wish to thank Alina Zare and Sheng Zou for providing the
NCM sampling code. We wish to acknowledge funding from the NASA Earth
and Space Science Fellowship Program and the Belgian Science Policy
Office in the framework of the STEREO III Program\textemdash{} Project
UrbanEARS (SR/00/307). We also thank the Jet Propulsion Laboratory
for providing radiometrically calibrated, orthorectified AVIRIS imagery. 

\bibliographystyle{11C__Users_Yuan_Zhou_Dropbox_YuanHyperspectral_document_GMM_SantaBarbara_IEEEbib1}

\begin{IEEEbiography}[{\includegraphics[width=1in,height=1.25in]{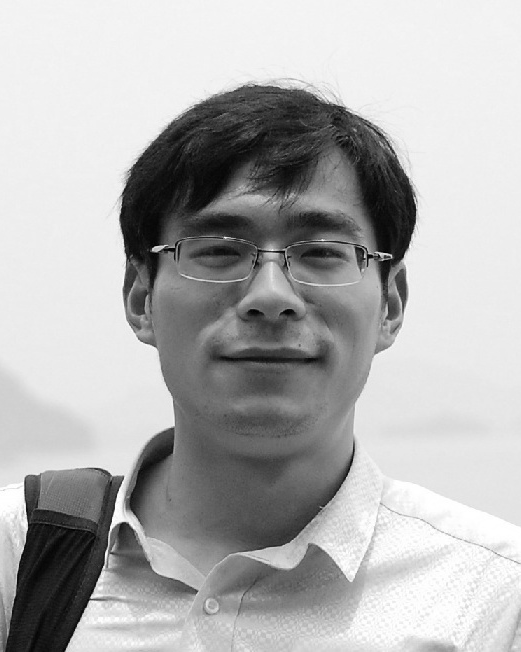}}]{Yuan Zhou}
 received the B.E degree in Software Engineering (2008), the M.E.
degree in Computer Application Technology (2011), both from Huazhong
University of Science and Technology, Wuhan, Hubei, China. Then he
worked in Shanghai UIH as a software engineer for two years. Since
2013, he has been a Ph.D. student in the Department of CISE, University
of Florida, Gainesville, FL, USA. His research interests include image
processing, computer vision and machine learning.
\end{IEEEbiography}

\begin{IEEEbiography}[{\includegraphics[width=1in,height=1.25in]{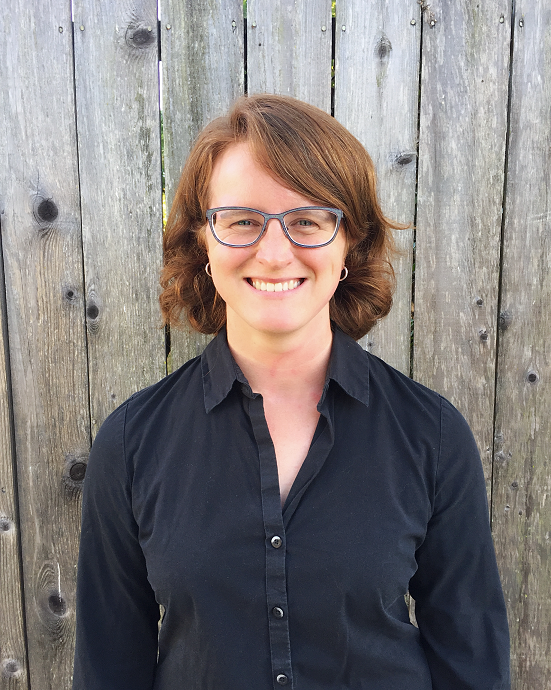}}]{Erin Wetherley}
 is a doctoral candidate in the Geography Department at the University
of California, Santa Barbara. She specializes in characterizing urban
environments using hyperspectral imagery, thermal imagery, and sub-pixel
analyses. Prior to her doctoral work, Erin earned a bachelors degree
in Environmental Studies from Brown University, and worked for several
years as a GIS and database manager at a Washington, D.C. non-profit
organization.
\end{IEEEbiography}

\begin{IEEEbiography}[{\includegraphics[width=1in,height=1.25in]{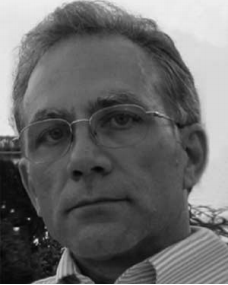}}]{Paul Gader}
 (M\textquoteright 86\textendash SM\textquoteright 09\textendash F\textquoteright 11)
received the Ph.D. degree in mathematics for image-processing-related
research from the University of Florida, Gainesville, FL, USA, in
1986. He was a Senior Research Scientist with Honeywell, a Research
Engineer and a Manager with the Environmental Research Institute of
Michigan, Ann Arbor, MI, USA, and a Faculty Member with the University
of Wisconsin, Oshkosh, WI, USA, the University of Missouri, Columbia,
MO, USA, and the University of Florida, FL, USA, where he is currently
a Professor of Computer and Information Science and Engineering. He
performed his first research in image processing in 1984 working on
algorithms for the detection of bridges in forward-looking infrared
imagery as a Summer Student Fellow at Eglin Air Force Base. He has
since worked on a wide variety of theoretical and applied research
problems including fast computing with linear algebra, mathematical
morphology, fuzzy sets, Bayesian methods, handwriting recognition,
automatic target recognition, biomedical image analysis, landmine
detection, human geography, and hyperspectral and light detection,
and ranging image analysis projects. He has authored/co-authored hundreds
of refereed journal and conference papers.
\end{IEEEbiography}

\end{document}